\documentclass[runningheads]{llncs}
\usepackage{graphicx}
\usepackage{comment}
\usepackage{amsmath,amssymb} 
\usepackage{color}

\usepackage{eso-pic}
\usepackage{xspace}

\makeatletter
\DeclareRobustCommand\onedot{\futurelet\@let@token\@onedot}
\def\@onedot{\ifx\@let@token.\else.\null\fi\xspace}

\def\eg{\emph{e.g}\onedot} 
\def\ie{\emph{i.e}\onedot} 
 
\def\etc{\emph{etc}\onedot} \def\vs{\emph{vs}\onedot}
 
\def\etal{\emph{et al}\onedot}
\makeatother

\usepackage{gensymb}
\usepackage{booktabs}

\usepackage{wrapfig}
\usepackage[font=footnotesize]{caption}
\usepackage{floatrow}
\usepackage{url}

\usepackage{subfig}

\usepackage{pifont}
\newcommand{\cmark}{\ding{52}}%
\newcommand{\xmark}{\ding{56}}%

\newcommand{\E}{\mathbb{E}}

\newcommand{\LL}{\mathcal{L}}
\newcommand{\WR}{\mathit{WR}}

\usepackage{etoolbox}
\newcommand{\zerodisplayskips}{%
  \setlength{\abovedisplayskip}{1ex}%
  \setlength{\belowdisplayskip}{1ex}%
  \setlength{\abovedisplayshortskip}{0ex}%
  \setlength{\belowdisplayshortskip}{0ex}}
\appto{\normalsize}{\zerodisplayskips}
\appto{\small}{\zerodisplayskips}
\appto{\footnotesize}{\zerodisplayskips}

\begin{document}
\pagestyle{headings}
\mainmatter
\def\ECCVSubNumber{4416}  

\title{Password-conditioned Anonymization and Deanonymization with Face Identity Transformers} 

\titlerunning{Password-conditioned Face Identity Transformers}
%
\author{Xiuye~Gu\inst{1,2}\orcidID{0000-0001-5568-564X} \and
Weixin~Luo\inst{2,3}\orcidID{0000-0002-0754-6458} \and
Michael~S.~Ryoo\inst{4}\orcidID{0000-0002-5452-8332} \and
Yong~Jae~Lee\inst{2}\orcidID{0000-0001-9863-1270}}

%
\authorrunning{X. Gu et al.}
%
\institute{$^1$Stanford University ~~ $^2$UC Davis ~~ $^3$ShanghaiTech ~~ $^4$Stony Brook University}
\maketitle

\begin{abstract}
Cameras are prevalent in our daily lives, and enable many useful systems built upon computer vision technologies such as smart cameras and home robots for service applications.  However, there is also an increasing societal concern as the captured images/videos may contain privacy-sensitive information (\eg,~face identity). We propose a novel \emph{face identity transformer} which enables automated photo-realistic password-based anonymization and deanonymization of human faces appearing in visual data. Our face identity transformer is trained to (1) remove face identity information after anonymization, (2) recover the original face when given the correct password, and (3) return a wrong---but photo-realistic---face given a wrong password. With our carefully designed password scheme and multi-task learning objective, we achieve both anonymization and deanonymization using the same single network. Extensive experiments show that our method enables multimodal password conditioned anonymizations and deanonymizations, without sacrificing privacy compared to existing anonymization methods.

\end{abstract}

\section{Introduction}\label{sec:introduction}
As computer vision technology is becoming more integrated into our daily lives, addressing privacy and security questions is becoming more important than ever.  For example, smart cameras and robots in homes are widely being used, but their recorded videos often contain sensitive information of their users.  In the worst case, a hacker could intrude these devices and gain access to private information. 

Recent anonymization techniques aim to alleviate such privacy concerns by redacting privacy-sensitive data like face identity information.  Some methods~\cite{butler15,ryoo2017privacy} perform low-level image processing such as extreme downsampling, image masking, etc.  A recent paper proposes to \emph{learn} a face anonymizer that modifies the identity of a face while preserving activity relevant information~\cite{jason}.  However, none of these techniques consider the fact that the video/image owner (and his/her friends, family, law enforcement, etc.) may want to see the \emph{original} identities and not the anonymized ones.  For example, people may not want their real faces to be saved directly on home security cameras due to privacy concerns; however, remote family members may want to see the real faces from time to time. Or when crimes arise, to catch criminals, police need to see their real faces.

This problem poses an interesting tradeoff between privacy and accessibility.  On the one hand, we would like a system that can anonymize sensitive regions (face identity) so that even if a hacker were to gain access to such data, they would not be able to know who the person is (without additional identity revealing meta-data).  On the other hand, the owner of the visual data inherently wants to see the original data, not the anonymized one.

\begin{figure}[t!]
\centering
\includegraphics[width=0.99\linewidth]{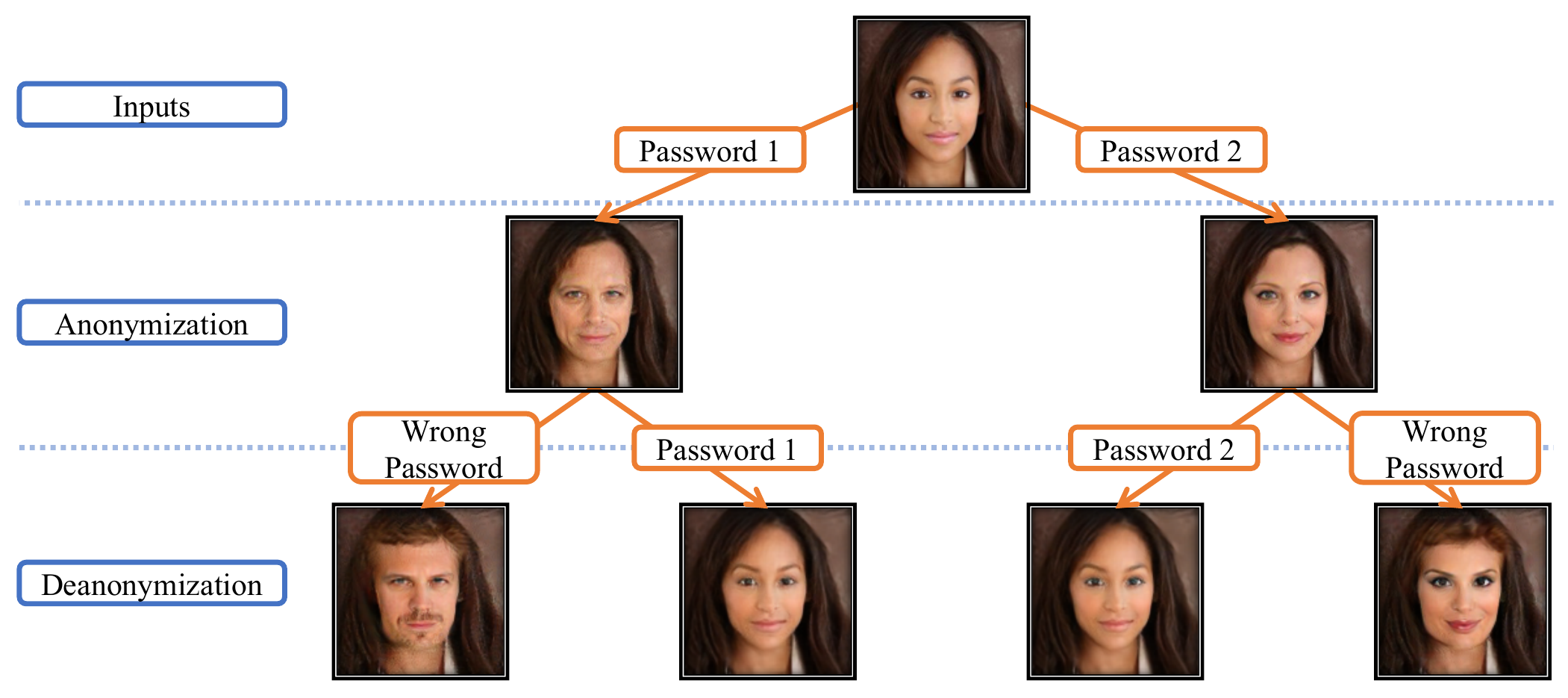}
\caption{Our system never stores users' faces on disk, and instead only stores the anonymized faces. When a user provides a correct recovery password, s/he will get the deanonymized face back. If a hacker invading their privacy inputs a wrong password, s/he will get a face whose identity is different from the original as well as the anonymized face. The photo-realism of the modified faces is meant to fool the hacker by providing no clues as to whether the real face was recovered.}   
\label{fig:teaser}
\end{figure}

To address this issue, we introduce a novel \emph{face identity transformer} that can both \emph{anonymize and deanonymize (recover)} the original image, while maintaining privacy.  We design a discrete password space, in which the password conditions the identity change. Specifically, given an original face, our face identity transformer outputs different anonymized face images with different passwords (Fig.~\ref{fig:teaser} Anonymization).  Then, given an anonymized face, the original face is recovered only if the correct password is provided (Fig.~\ref{fig:teaser} Deanonymization, `Password 1/2').  We further increase security as follows: Given an anonymized face, if a wrong password is provided, then it changes to a new identity, which is still different from the original identity  (Fig.~\ref{fig:teaser} Deanonymization, `Wrong Password').  Moreover, each wrong password maps to a unique identity.  In this way, we provide security via ambiguity: even if a hacker guesses the correct password, it is extremely difficult to know that without having access to any other identity revealing meta-data, since each password---regardless of whether it is correct or not---always leads to a different realistic identity.

To enforce the face identity transformer to output different anonymized face identities with different passwords, we optimize a multi-task learning objective, which includes maximizing the feature-level dissimilarity between pairs of anonymized faces that have different passwords and fooling a face classifier. To enforce it to recover the original face with the correct password, we train it to anonymize and then recover the correct identity only when given the correct password, and to produce a new identity otherwise. Lastly, we maximize the feature dissimilarity between an anonymized face and its deanonymized face with a wrong password so that the identity always changes. Moreover, considering the limited memory space on devices, we propose to use the same single transformer to serve both anonymization and deanonymization purposes.

We note that our approach is related to cryptosystems like RSA~\cite{rivest1978method}.  The key difference is that cryptosystems do not produce encryptions that are visually recognizable to human eyes.  However, in various scenarios, users may want to understand what is happening in anonymized visual data.  For example, people may share photos/videos over public social media with anonymized faces, but only their real-life friends have the passwords and can see their real faces to protect identity information. Moreover, with photorealistic anonymizations, one can \emph{easily apply existing computer vision based recognition algorithms on the anonymized images} as we demonstrate in Sec.~\ref{sec:cvalgorithm}. In this way, it could work with e.g., smart cameras that use CV algorithms to analyze content but in a privacy-preserving way, unlike other schemes (\eg, homomorphic encryption) that require developing new ad-hoc recognition methods specific to nonphotorealistic modifications, in which accuracy may suffer.

In our approach, only the anonymized data is saved to disk (\ie, conceptually, the anonymization would happen at the hardware-level via an embedded chipset -- the actual implementation of which is outside the scope of this work). The advantage of this concept is that the hacker could never have direct access to the original data. Finally, although there may be other identity-revealing information such as gait, clothing, background, etc., our work entirely focuses on improving privacy of face identity information, but would be complementary to systems that focus on those other aspects.

Our experiments on CASIA \cite{CASIA}, LFW \cite{LFW2}, and FFHQ \cite{karras2018style} show that the proposed method enables multimodal face anonymization as well as recovery of original face images, without sacrificing privacy compared to existing advanced anonymization~\cite{jason} and classical image processing techniques including masking, noising, and blurring, etc. \emph{Please see} \url{https://youtu.be/FrYmf-CL4yk} \emph{and Fig.~6 in the supp for image/video in the wild results.}

\section{Related work}
\paragraph{Privacy-preserving visual recognition.}
This is the problem of detecting humans, their actions, and objects without accessing user-sensitive information in images/videos.  Some methods employ extreme low-resolution downsampling to hide sensitive details~\cite{wang2016studying,chen2017semi,ryoo2017privacy,ryoo2018extreme} but suffer from lower recognition performance in downstream tasks. More recent work propose a head inpainting obfuscation technique~\cite{inpainting}, a four-stage pipeline that first obfuscates facial attributes and then synthesizes faces~\cite{Anonymousnet}, and a video anonymizer that performs pixel-level modifications to remove people's identity while preserving motion and object information for activity detection~\cite{jason}. Unlike our approach, none of these existing work employ a password scheme to condition the anonymization, and also do not perform deanonymization to recover the original face.  Moreover, even if one could brute-forcely train a deanonymizer for these methods, there is no way to provide wrong recoveries upon wrong passwords, as our method does. 

Security/cryptography research on privacy-preserving recognition is also related \eg,~\cite{erkin2009privacy,gilad2016cryptonets}. The key difference is that these methods encrypt data in a secure but visually-uninterpretable way, whereas our goal is to anonymize the data in a way that is still interpretable to humans and existing computer vision techniques can still be applied. Differential privacy~\cite{abadi2016privacy,yonetani2017privacy} is also related but its focus is on protecting privacy in the training data whereas ours is on anonymizing visual data during the inference stage. 

\paragraph{Face image manipulation and conditional GANs.}
Our work builds upon advances in pixel-level synthesis and editing of realistic human faces~\cite{larsen2015autoencoding,perarnau2016invertible,shen2017learning,karras2018style,ganimation,IPGAN} and conditional GANs~\cite{mirza2014conditional,reed2016generative,pix2pix,zhang2017stackgan,zhu2017your,cao2018hashgan,infogan,Regmi_2018_CVPR}, but we differ significantly in our goal, which is to completely change the identity of a face (and also recover the original) for privacy-preserving visual recognition.


\begin{figure*}[t!]
\centering
\includegraphics[width=\linewidth]{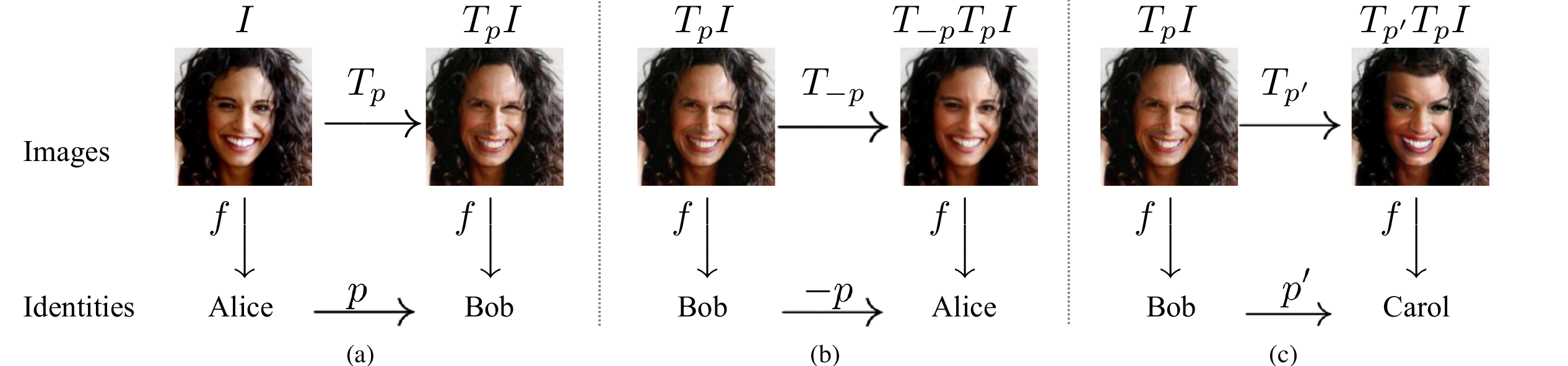}
   \caption{Privacy-preserving properties that our face identity transformer $T$ learns. (a) Anonymization stage. (b) Deanonymization stage with correct recovery password. (c) Deanonymization stage with incorrect recovery password.}
\label{fig:equivariant}
\end{figure*}

\section{Desiderata} \label{sec:formulation}
Our face identity transformer $T$ takes as input a face image $I \in \Phi$ and a user-defined password $p \in P$, where $\Phi$ and $P$ denote the face image domain and password domain.  We use the notation $T_pI$ to denote the transformed image with input image $I$ and password $p$. Before diving into the details, we first outline desired properties of a privacy-preserving face identity transformer.

\paragraph{Minimal memory consumption.}  Considering the limited memory space on most camera systems, a \textit{single} face identity transformer that can both anonymize and deanonymize faces is desirable.

\paragraph{Photo-realism.}  
We would like the transformer to maintain photo-realism for any transformed face image:
\begin{align}
T_pI \in \Phi, \quad \forall p \in P, \forall I \in \Phi. \label{eq:photo_realistic_condition}
\end{align}

Photo-realism has three benefits: 1) a human who views the transformed images will still be able to interpret them; 2) one can easily apply existing computer vision algorithms on the transformed images; and 3) it's possible to confuse a hacker, since photo-realism can no longer be used as a cue to differentiate the original face from an anonymized one.   

\paragraph{Compatibility with background.} 
The background $B(\cdot)$ of the transformed face should be the same as the original:
\begin{align}
B(T_pI) = B(I),\quad \forall p\in P, \forall I \in \Phi. \label{eq:background}
\end{align}

This will ensure that there are no artifacts between the face region and the rest of the image (\ie, it will not be obvious that the image has been altered).

\paragraph{Anonymization with passwords.} 
Let $f:\Phi \rightarrow \Gamma$ denote the function mapping face images to people's identities.
We would like to condition anonymization via a password $p$:
 \begin{align}
f(T_pI) \neq f(I), \quad \forall p \in P, \forall I \in \Phi. \label{eq: anonymization_condition}	
\end{align}

\paragraph{Deanonymization with inverse passwords.} 
We should recover the original identity only when the correct password is provided. To achieve our goal of minimal memory consumption, we can model the additive inverse of the password used for anonymization as the correct password for deanonymization.  In this way, we can use the same transformer for deanonymization, \ie~we model $T_{-p} = T_p^{-1}$:
\begin{align}
f(T_{-p}T_pI) = f({T}^{-1}_{p} T_pI)  = f(I),\ \forall p \in P, \forall I \in \Phi. \label{eq: de_anonymization_condition_recover}
\end{align}

\paragraph{Wrong deanonymization with wrong inverse passwords.} 
We would like the transformer to change the anonymized identity into a \textit{different} identity that is different from both the original as well as the anonymized image when given a wrong inverse password:
\begin{align}
f(T_{p'} T_p I ) \neq f(I),\quad \forall p,p' \in P, p' \neq -p, \forall I \in \Phi, \label{eq: de_anonymization_condition_wrong_not_original}\\
f(T_{p'} T_p I ) \neq f(T_pI),\ \forall p,p' \in P, p' \neq -p, \forall I \in \Phi. \label{eq: de_anonymization_condition_wrong_not_fake}
\end{align}

In this way, whether the password is correct or not, the identity is always changed so as to confuse the hacker.

\paragraph{Diversity.}
The image $I$ should be transformed to different identities with different passwords, to increase security in both anonymization and deanonymization. Otherwise, if multiple passwords produce the same identity, a hacker could realize that the photo is anonymized or his attempts have failed in deanonymization:
\begin{align}
f(T_{p_1}I) \neq f(T_{p_2}I),\ \forall p_1, p_2 \in P, p_1 \neq p_2, \forall I \in \Phi. \label{eq:diversity}
\end{align}

\noindent Fig.~\ref{fig:equivariant} summarizes our desiderata for anonymization and deanonymization. 

\begin{figure*}[t!]
\centering
\includegraphics[width=\linewidth]{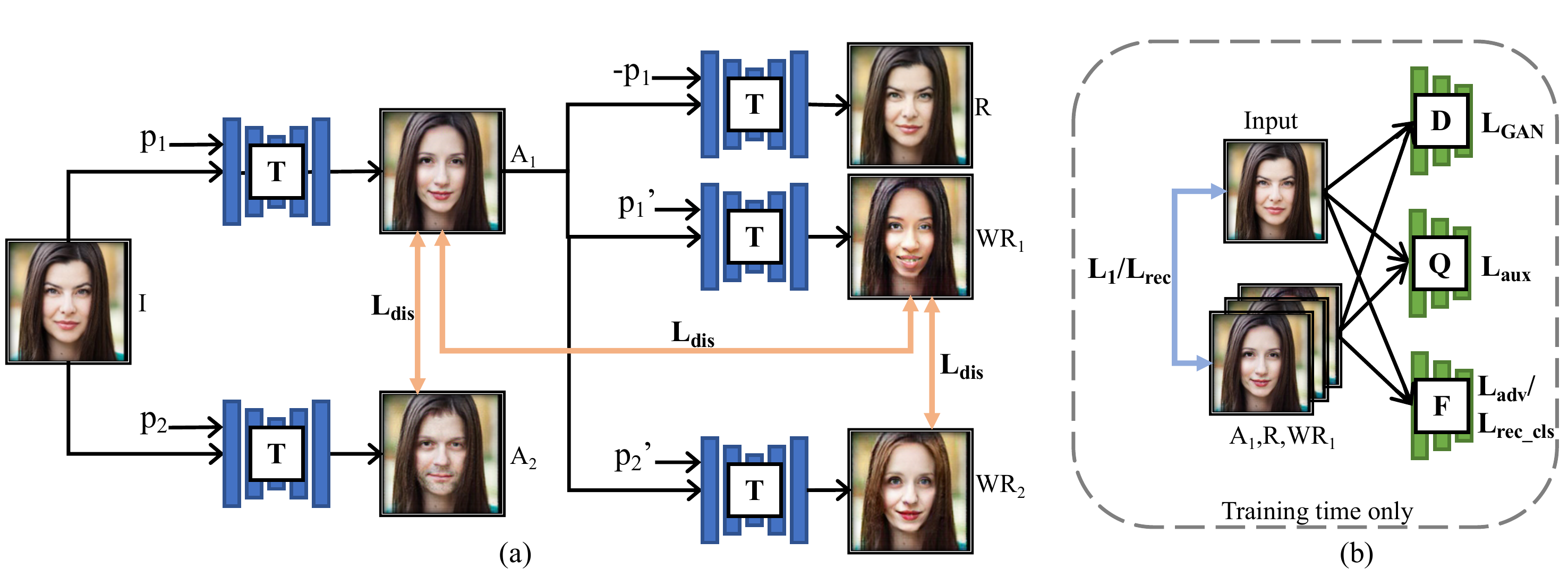}
   \caption{(a) Face identity transformer network architecture. (b) Objectives we apply to synthesized images during training (not included in (a) for clarity). $I$: Input image, $A_{1,2}$: Anonymized faces, $R$: Recovered face, $\WR_{1,2}$: Wrongly Recovered faces. $\LL_{feat}$ is the sum of the three $\LL_{dis}$'s.}
\label{fig:arch}
\end{figure*}

\section{Approach: \textit{Face Identity Transformer}}
Our face identity transformer $T$ is a conditional GAN trained with a multi-task learning objective.  It is conditioned on both the input image $I$ and an input password $p$.  Importantly, the function of $p$ is different from the usual random noise vector $z$ in GANs: $z$ simply tries to model the distribution of the input data, while $p$ in our case makes the transformer hold the desired privacy-preserving properties (Eq.~\ref{eq: anonymization_condition}-\ref{eq:diversity}). We next explain our password scheme, multimodal identity generation, and multi-task learning objective.

\subsection{Password scheme} \label{subsec: password_embedding}
We use an $N$-bit string $p \in \{0, 1\}^N$ as our password format, which consists of $2^N$ unique passwords. Given image $I \in \mathbb{R}^{H \times W \times 3}$, we form the input to the transformer as a depthwise concatenation $(I, p) \in \mathbb{R}^{H \times W \times (3+N)}$, where $p$ is replicated in every pixel location. To make the transformer condition its identity change on the input password, we design an auxiliary network $Q(I, T_pI) = \hat{p}$. It learns to predict the embedded password from the input and transformed image pair, and thus maximizes the mutual information between the injected password and the identity change in the image domain, similar to InfoGAN~\cite{infogan}. We use cross entropy loss for the classifier $Q$, and denote it as $\LL_{aux}(T, Q)$. See supp Sec.~1 for the detailed formula.



\subsection{Multimodal identity change} \label{subsec:feature_dismatching}
Conditional GANs with random noise do not produce highly stochastic outputs~\cite{Mathieu_et_al,pix2pix}. To overcome this, BicycleGAN~\cite{multimodal} uses an explicitly-encoded multimodality strategy similar to our auxiliary network $Q$. However, even with $Q$, we only observe multimodality on colors and textures as in~\cite{multimodal}, but not on high-level face identity.

Thus, to induce diverse high-level identity changes, we propose an explicit feature dissimilarity loss.
Specifically, we use a face recognition model $F$ to extract deep embeddings of the faces, and minimize their cosine similarity when they are associated with different passwords:
\begin{align}
	\LL_{dis}(M_1, M_2) = \max\Big(0, \cos\big(F_{embed}(M_1), F_{embed}(M_2)\big)\Big), 
\end{align}
where $\cos$ is cosine similarity, and $M_1$ and $M_2$ are two transformed face images with two different passwords.  We do not penalize pairs whose cosine similarity is less than 0; \ie, it is enough for the faces to be different up to a certain point.

We apply the dissimilarly loss between (1) two anonymized faces with different passwords, (2) two incorrectly deanonymized faces given different wrong passwords, and (3) the anonymized face and wrongly recovered face:
\begin{align}
	\LL_{feat}(T) &= \E_{(I,p_1 \neq p_2)} \LL_{dis}(T_{p_1}I, T_{p_2}I)\nonumber\\
	&+ \E_{(I,p_1' \neq p_2', p_1' \neq -p, p_2' \neq -p)} \LL_{dis}(T_{p_1'}T_pI, T_{p_2'}T_pI)\nonumber\\
	&+ \E_{(I,p' \neq -p)} \LL_{dis}(T_{p}I, T_{p'}T_pI).
\end{align}

This loss can be easily satisfied when the model outputs extremely different content that do not necessarily look like a face, and thus can adversely affect other desideratum (\eg, photo-realism) of a privacy-preserving face identity transformer. We next introduce a multi-task learning objective to restrict the outputs to lie on the face manifold, as a form of regularization.

\subsection{Multi-task learning objective}
We describe our multi-task objective that further aids identity change, identity recovery, and photo-realism.

\paragraph{Face classification adversarial loss.} We apply the face classification adversarial loss from~\cite{jason}, which helps change the input face's identity.  We apply it on both the transformed face $T_pI$ as well as the reconstructed face with wrong recovery password $T_{p'}T_pI$:
\begin{align}
\LL_{adv}(T,F) = &-\E_I \LL_{CE}\big(F(I), y_I\big) -\E_{(I,p)} \LL_{CE}\big(F(T_pI), y_I\big) \nonumber\\
 &- \E_{(I,p'\neq -p)} \LL_{CE}\big(F(T_{p'}T_pI), y_I\big),	
\end{align}
where $F$ is the face classifier, $y_I$ is face identity label, and $\LL_{CE}$ denotes cross entropy loss.

Similar to the dissimilarity loss ($\LL_{dis}$), this loss pushes the transformed face to have a different identity.  The key difference is that this loss requires face identity labels so cannot be used to push $T_{p_1}I$ and $T_{p_2}I$ to have different identities, but has the advantage of utilizing supervised learning so that it can change the identity more directly.

\paragraph{Reconstruction losses.} We use $L_1$ reconstruction loss for deanonymization:
\begin{align}
	\LL_{rec}(T) = \|T_{-p}T_pI - I \|_1.
\end{align}

With the $L_1$ loss alone, we find the reconstruction to be often blurry. Hence, we also introduce a face classification loss $\LL_{rec\_cls}$ on the reconstructed face to enforce the transformer to recover the high-frequency identity information:
\begin{align}
\LL_{rec\_cls}(T,F) = \E_{(I,p)} \LL_{CE}\big(F(T_{-p}T_pI), y_I\big). 
\end{align}

This loss enforces the reconstructed face $T_{-p}T_pI$ to be predicted as having the same identity as $I$ by face classifier $F$.

\paragraph{Background preservation loss.} For any transformed face, we try to preserve its original background.  To this end, we apply another $L_1$ loss (with lower weight):
\begin{align}
\LL_{1}(T) = \|T_pI - I\|_1 + \|T_{p'}T_pI - I\|_1.
\end{align}

Although employing a face segmentation algorithm is an option, we find that applying $\LL_1$ on the whole image works well to preserve the background.

\paragraph{Photo-realism loss.} 
We use a photo-realism adversarial loss $\LL_{GAN}$~\cite{GAN} on generated images to help model the distribution of real faces. Specifically, we use PatchGAN~\cite{pix2pix} to restrict the discriminator $D$'s attention to the structure in local image patches. To stabilize training, we use LSGAN~\cite{lsgan}:

\begin{align}
\max_{D}\mathcal{L}_{GAN}(D) &= -\frac{1}{2} \mathbb{E}_I [\big(D(I) - 1\big)^2] - \frac{1}{2} \mathbb{E}_{(I, p)} [ D(T_pI)^2 ]\\
\min_{T}\mathcal{L}_{GAN}(T) &= \mathbb{E}_{(I, p)} [ \big(D(T_pI) - 1\big)^2 ]	
\end{align}

\subsection{Full objective}
Overall, our full objective is:
\begin{align}
	\LL &= \lambda_{aux} \LL_{aux}(T, Q) + \lambda_{feat} \LL_{feat}(T)\nonumber\\
	&+ \lambda_{adv} \LL_{adv}(T, F) + \lambda_{rec\_cls} \LL_{rec\_cls}(T, F)  \nonumber \\
	&+ \lambda_{rec} \LL_{rec}(T) + \lambda_{L_1} \LL_1(T) + \LL_{GAN}(T, D). \label{eq: full_objective}
\end{align}
We optimize the following minimax problem to obtain our face identity transformer:
\begin{align}
	T^* = \arg \min_{T, Q} \max_{D, F} \LL \label{eq:minimax}	
\end{align}

\paragraph{Training.} Fig.~\ref{fig:arch} shows our network for training. For each input $I$, we randomly sample two different passwords for anonymization and two incorrect passwords for wrong recoveries, and then impose $\LL_{dis}$ on the generated pairs and enforce $\LL_{dis}$ between the anonymization and wrong reconstruction. We observe that during training, the auxiliary networks and backprop can consume a lot of GPU memory, which limits batch size.  We propose a strategy based on symmetry: except for the feature dissimilarity loss, we apply all other losses only to the first anonymization and first wrong recovery, which empirically works well. 

We adopt a two-stage training strategy for the minimax problem~\cite{GAN}. In the discriminator's stage, we fix the parameters of $T,Q$, and update $D,F$; in the generator's stage, we fix $D,F$, and update $T,Q$.

\paragraph{Inference.} During testing, the transformer $T$ takes as input a user-defined password and a face image, anonymizes the face, and saves it to disk. When the user/hacker wants to see the original image, the transformer takes the recovery password and the anonymized image, and either outputs the identity-recovered image or a hacker-fooling image depending on password correctness. Throughout the whole process, the original images and passwords are never saved on disk for privacy reasons.

\section{Experiments}
In this section, we demonstrate that our face identity transformer achieves password conditioned anonymization and deanonymization with photo-realism and multimodality. We also conduct ablation studies to analyze each module/loss.

\paragraph{Implementation details.}
Our identity transformer $T$ is built upon the network from~\cite{cyclegan}. We use size 128x128 for both inputs and outputs.
We subtract 0.5 from $p$ before inputting it to the transformer to make the password channels have zero mean. We set $N$=16. 
We use the pretrained SphereFace~\cite{sphereface} as our face recognition network $F$ for both deep embedding extraction in the feature dissimilarity loss and face classification adversarial training. 
For each stage, we use two PatchGAN discriminators~\cite{pix2pix} that have identical structure but operate at different image scales to improve photo-realism. The coarser discriminator is shared among all stages, while three separate finer discriminators are used for anonymization, reconstruction, and wrong recovery.
To improve stability, we use a buffer of 500 generated images when updating $D$.
We set $\lambda_{aux}$=1, $\lambda_{feat}$=2, $\lambda_{adv}$=2, $\lambda_{rec\_cls}$=1, $\lambda_{L_1}$=10 and $\lambda_{rec}$=100, based on qualitative observations.

\paragraph{Datasets.}
1) CASIA~\cite{CASIA} has 454,590 face images belonging to 10,574 identities. We split the dataset into training/validation/testing subsets made up of $80\%/10\%/10\%$ identities. We use the validation set to select our model. All reported results are on the test set.
2) LFW~\cite{LFW2} has 13,233 face images belonging to 5,749 identities. As our network is never trained on LFW, we evaluate on the entire LFW to test generalization ability. 
3) FFHQ~\cite{karras2018style} is a high-quality face dataset for benchmarking GANs. It is not a face recognition dataset, so we use it to only test generalization. We directly test our model on its validation set at 128x128 resolution, which contains 10,000 images.

\paragraph{Evaluation metrics.}
\textbf{\emph{Face verification accuracy:}} We measure our transformer's identity changing ability with a standard binary face verification test, which scores whether a pair of images have the same identity or not. Since different face recognition models may have different biases, we use two popular pretrained face recognition models: SphereFace~\cite{sphereface} and VGGFace2~\cite{vggface2}.

\noindent\textbf{\emph{Face recovery quality:}} We measure face recovery quality using \textbf{LPIPS distance}~\cite{lpips}, which measures perceptual similarity between two images based on deep features, and \textbf{DSSIM}~\cite{DSSIM}, which is a commonly-used low-level perceptual metric. We also use pixel-level $L_1$ and $L_2$ distance.

\noindent\textbf{\emph{AMT perceptual studies:}} We use Amazon Mechanical Turk (AMT) to test how well our method 1) changes and recovers identities, 2) achieves photo-realism, and 3) attains multimodal anonymizations, as judged by human raters. 

\noindent\textbf{\emph{Runtime:}} On a single Titan V, averaged over CASIA testset, runtime is 0.0266 sec/batch with 12 images per batch. Though we use multiple auxiliary networks to help achieve our desiderata, they are all discarded during inference time.

\begin{table}[t!]
\centering
\scriptsize
\begin{tabular}{lccc}
\hline
Method & Anonymize? & Deanonymize? & Password-conditioned?  \\
\hline
Ren \etal~\cite{jason} & \cmark & \cmark & \xmark\\
Super-pixel & \cmark & \cmark & \xmark\\
Edge & \cmark & \cmark & \xmark\\
Blur & \cmark & \xmark & \xmark\\
Noise & \cmark & \cmark & \xmark\\
Masked & \cmark & \xmark & \xmark\\
\textbf{Ours} & \cmark & \cmark & \cmark\\
\hline
\end{tabular}
\caption{Privacy-preserving ability comparison. Our method is the only one that supports password-conditioned face (de)anonymization without sacrificing privacy.} \label{table:three_anonymization_properties}
\end{table}

\subsection{Anonymization and deanonymization}\label{subsec:ano_deano}

To our knowledge, \emph{no prior work achieves password-conditioned anonymization and deanonymization on visual data like ours}, see Table~\ref{table:three_anonymization_properties}. Hence, we cannot directly compare with any existing method on generating \emph{multimodal} anonymizations and deanonymizations.

Despite this, we want to ensure that our method does no worse than existing methods in terms of anonymization and deanonymization (setting aside the password conditioning capability).  To demonstrate this, following~\cite{jason}, we compare to the following baselines:
\textbf{Ren \etal}~\cite{jason}: a learned face anonymizer that maintains action detection accuracy;
\textbf{Superpixel}~\cite{butler15}: each pixel's RGB value is replaced with its superpixel's mean RGB value;
\textbf{Edge}~\cite{butler15}: face regions are replaced with corresponding edge maps;
\textbf{Blur}~\cite{ryoo2017privacy}: images are downsampled to extreme low-resolution ($8\times8$) and then upsampled back;
\textbf{Noise}: strong Gaussian noise ($\sigma^2 = 0.5$) is added to the image; 
\textbf{Masked}: face areas ($0.6\times$ of the face image) are masked out.

We also train deanonymizers for each baseline (\ie, to recover the original face), by using the same generator architecture with our reconstruction and photo-realism losses. Please refer to supp Fig.~1 for a qualitative example of the baselines and their anonymizations/deanonymizations.

\begin{figure*}[t]
\centering
\includegraphics[width=\linewidth]{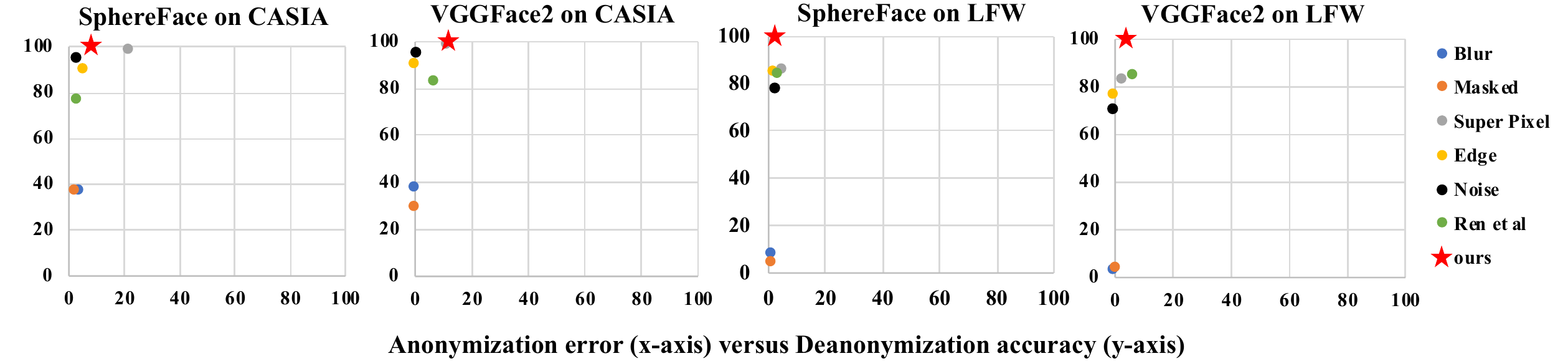}
   \caption{Anonymization vs.~deanonymization quality, measured by face verification error/accuracy on CASIA and LFW. Top-left corner is ideal. This result shows that we don't sacrifice (de)anonymization ability by introducing password conditioning.}
\label{fig:accuracy_plot}
\end{figure*}

Fig.~\ref{fig:accuracy_plot} shows anonymization vs. deanonymization (recovery) quality on CASIA and LFW using SphereFace and VGGFace2 as our face recognizers.  Our approach performs competitively to Ren \etal~\cite{jason}, ``Superpixel'', ``Edge'', ``Blur'' , ``Noise'', and ``Masked'' when considering both anonymization and deanonymization quality together.  This result confirms that we do not sacrifice the ability to anonymize/deanonymize by introducing password-conditioning.  In fact, in terms of reconstruction (deanonymization) quality (Table~\ref{table:reconstruction_quality}), our method outperforms the baselines by a large margin because we train our identity transformer to do anonymization and deanonymization in conjunction in an end-to-end way.

\begin{table}[t]
\centering
\scriptsize
\begin{tabular}{@{}l@{}|cccc|cccc@{}}
\toprule
& \multicolumn{4}{c|}{CASIA} & \multicolumn{4}{c}{LFW} \\
\midrule
Method & LPIPS & DSSIM & $L_1$ & $L_2$ & LPIPS & DSSIM & $L_1$ & $L_2$ \\
\midrule
Ren et al & 0.08 & 0.07 & 0.06 & 0.009 & 0.08 & 0.07 & 0.06 & 0.010 \\
Superpixel & 0.09 & 0.10 & 0.06 & 0.01 & 0.10 & 0.11 & 0.07 & 0.02 \\
Edge & 0.25 & 0.24 & 0.25 & 0.24 & 0.28 & 0.26 & 0.29 & 0.18  \\
Blur &  0.30 & 0.21 & 0.12 & 0.04 & 0.34 & 0.24 & 0.14 & 0.05 \\
Noise &  0.12 & 0.12 & 0.07 & 0.01 & 0.13 & 0.12 & 0.08 & 0.01\\
Masked & 0.10 & 0.09 & 0.07 & 0.02 & 0.16 & 0.13 & 0.10 & 0.05\\
Ours & \bfseries 0.03 & \bfseries 0.03 & \bfseries 0.04 & \bfseries 0.004 &\bfseries 0.04 & \bfseries 0.03 & \bfseries 0.04 & \bfseries 0.004\\
\bottomrule
\end{tabular}
\caption{CASIA and LFW reconstruction error. Ours produces best deanonymizations.} \label{table:reconstruction_quality}
\end{table}

Lastly, we perform AMT perceptual studies to rate our anonymizations and deanonymizations.  Specifically, we randomly sample 150 testing images ($I$), and generate for each image: an anonymized face with a random password ($A$), a recovered face with correct inverse password ($R$), and a recovered face with wrong password ($\WR$).  We then distribute 600 $I~vs~A$, $I~vs~R$, $I~vs~\WR$, and $A~vs~\WR$ pairs to turkers and ask ``Are they the same person?''. For each pair, we collect responses from 3 different turkers and take the majority as the answer to reduce noise.

The turkers reported $\mathbf{4.7}\%$ / $\mathbf{100}\%$ / $\mathbf{0.7}\%$ / $\mathbf{1.3}\%$ on $I vs A$ / $I vs R$ / $I vs \WR$ /$A vs \WR$. (low, high, low, low is ideal.) This further shows our method obtains the desired password-conditioned anonymization/deanonymization goals. We show all failure pairs for $I vs A$ in supp Sec.~5 and analyze the error there.
%

\subsection{Photo-realism}

To evaluate whether our (de)anonymization affects photo-realism, we conduct AMT user studies.  We follow the same perceptual study protocol from~\cite{cyclegan} and test on both anonymizations and wrong recoveries. For each test, we randomly generate 100 ``real \vs fake'' pairs. For each pair, we average responses from 10 unique turkers. Turkers label our anonymizations as being more real than a real face $\mathbf{28.9}\%$ of the time, and label our wrong reconstructions as more real than a real face $\mathbf{15.4}\%$  of the time. (Chance performace is $50\%$.) This shows that our generated images are quite photo-realistic.

\begin{figure*}[t!]
\centering
\includegraphics[width=\linewidth]{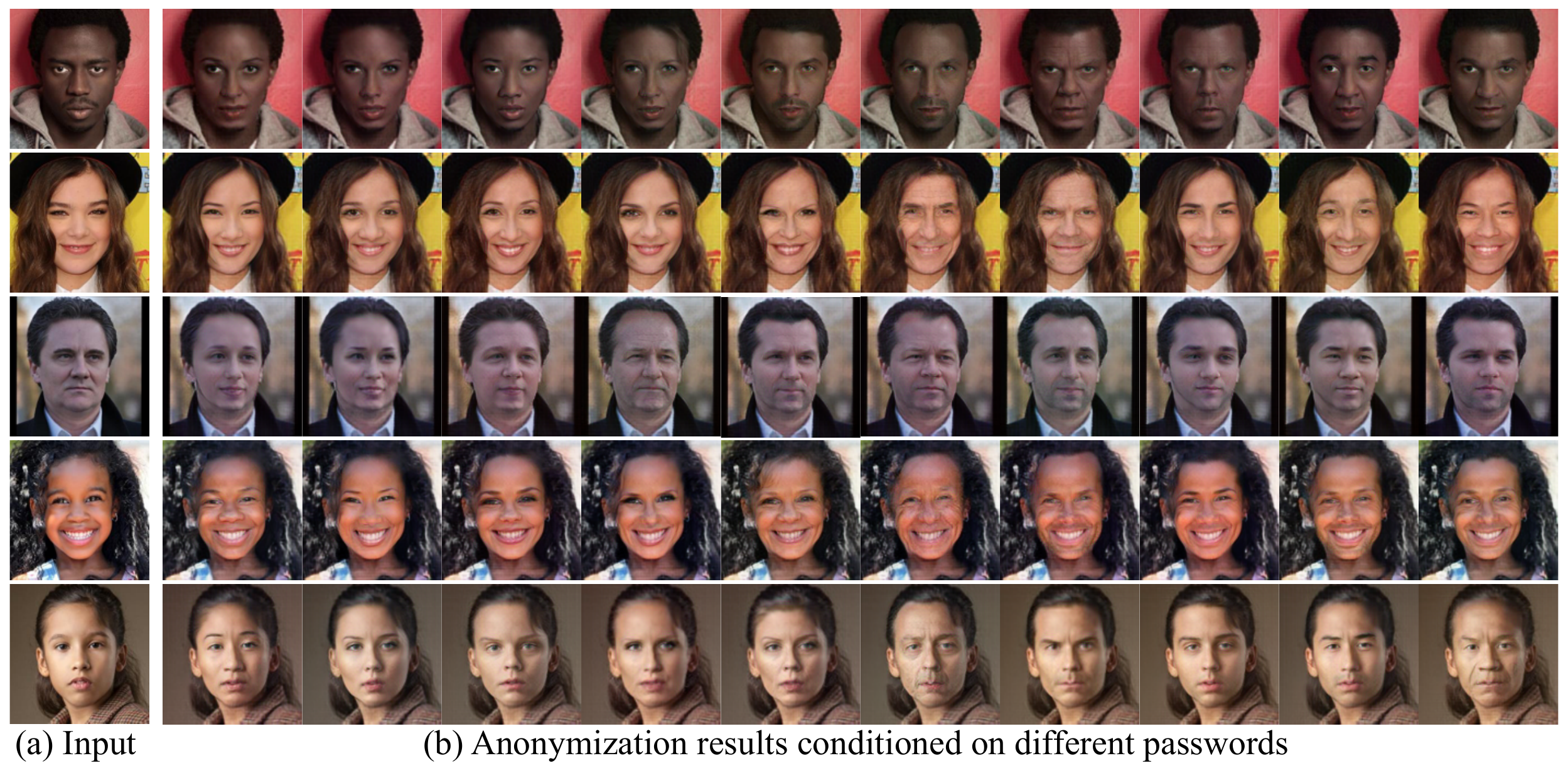}
   \caption{Multimodality results on CASIA. We observe a wide range of identity changes with different passwords.}
\label{fig:multimodal}
\end{figure*}

\subsection{Multimodality}
We next evaluate our model's ability to create different faces given different passwords.   Fig.~\ref{fig:multimodal} shows qualitative results. Our transformer successfully changes the identity into a broad spectrum of different identities, from women to men, from young to old, \etc.

We quantitatively evaluate multimodality through an AMT perceptual study.   We ask AMT workers to compare 150 $A_1~vs~A_2$ and 150 $\WR_1~vs~\WR_2$ pairs (pairs of anonymized / wrong-recovered faces with different passwords generated from the same input image) and ask ``are they the same person?". The turkers reported ``yes'' only $\mathbf{12.2\%}$ and $\mathbf{2.7\%}$ of the time, respectively (lower is better).  The results show that our transformer does quite well in generating different identities given different passwords.

\begin{figure}[t!]
\centering
\subfloat[\footnotesize{FFHQ}]{\includegraphics[width=0.48\linewidth]{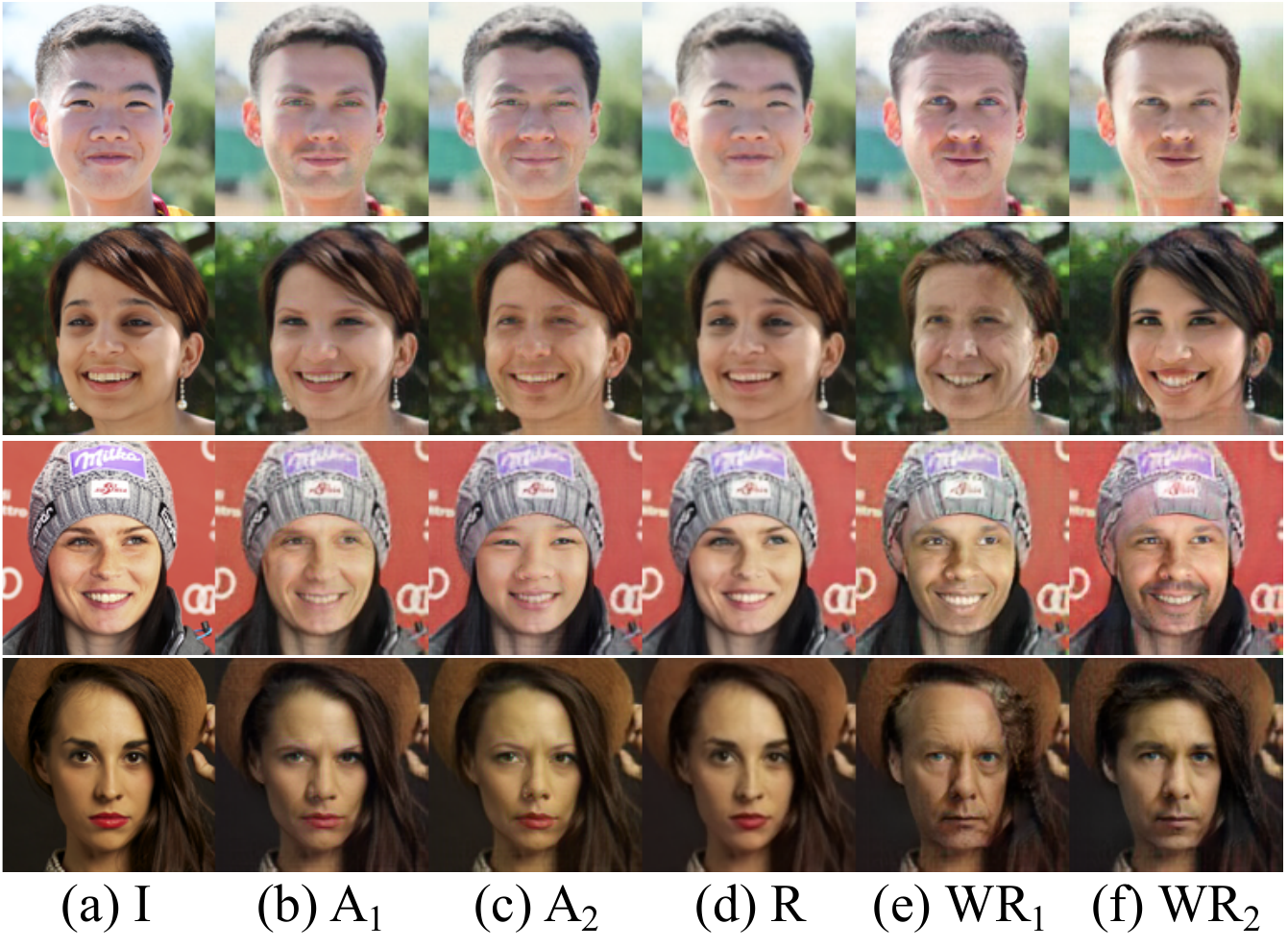}\label{fig:ffhq}}
\hfill
\subfloat[\footnotesize{LFW}]{\includegraphics[width=0.48\linewidth]{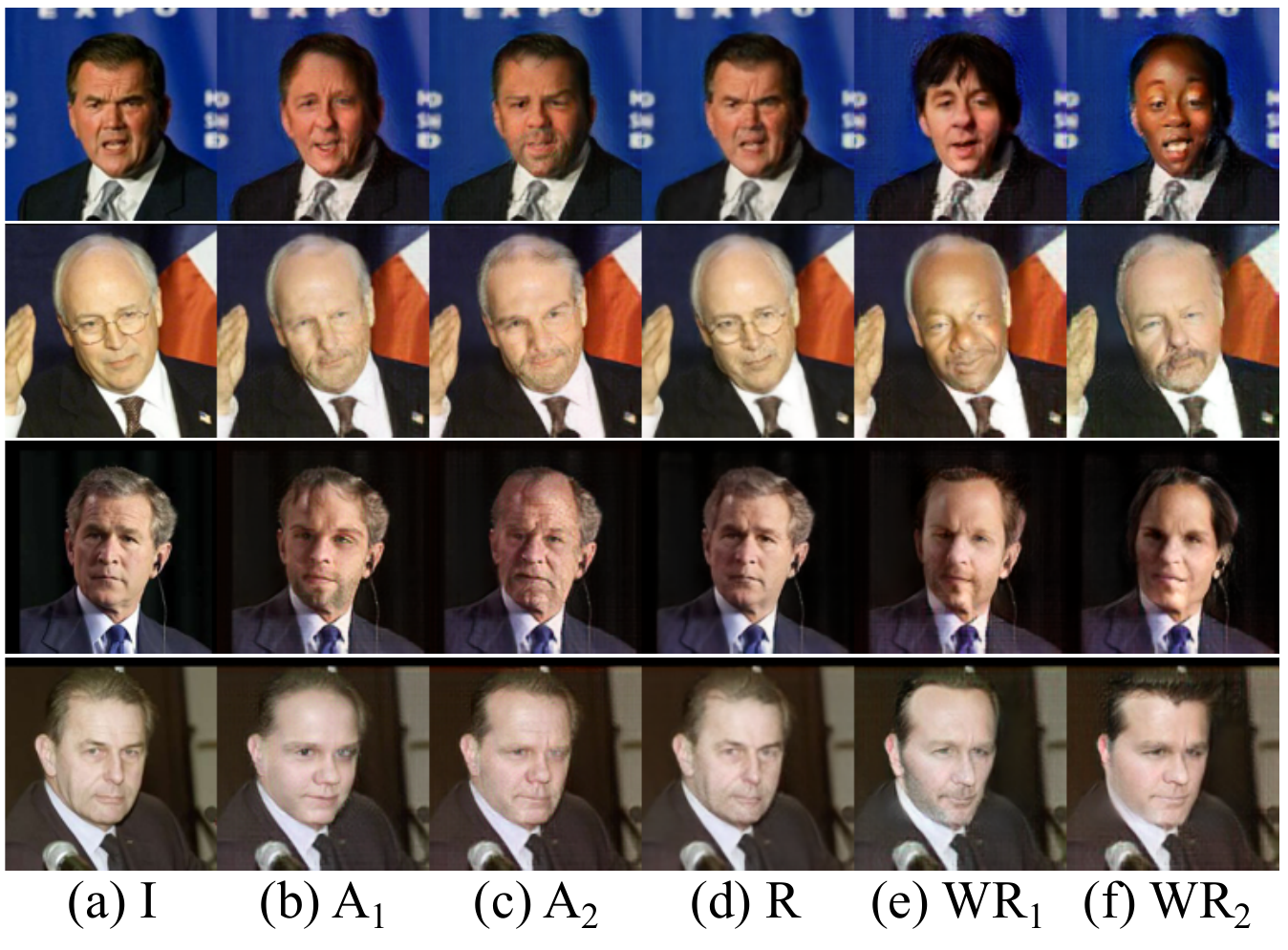}\label{fig:lfw}}
\caption{FFHQ and LFW generalization results. $I$: original image, $A_{1,2}$: anonymized faces using different passwords, $R/\WR_{1,2}$: recovered faces with correct/wrong passwords.}\label{fig:ffhq_and_lfw}
\end{figure}

\begin{figure}[t!]
\begin{floatrow}
\ffigbox{
\includegraphics[width=\linewidth]{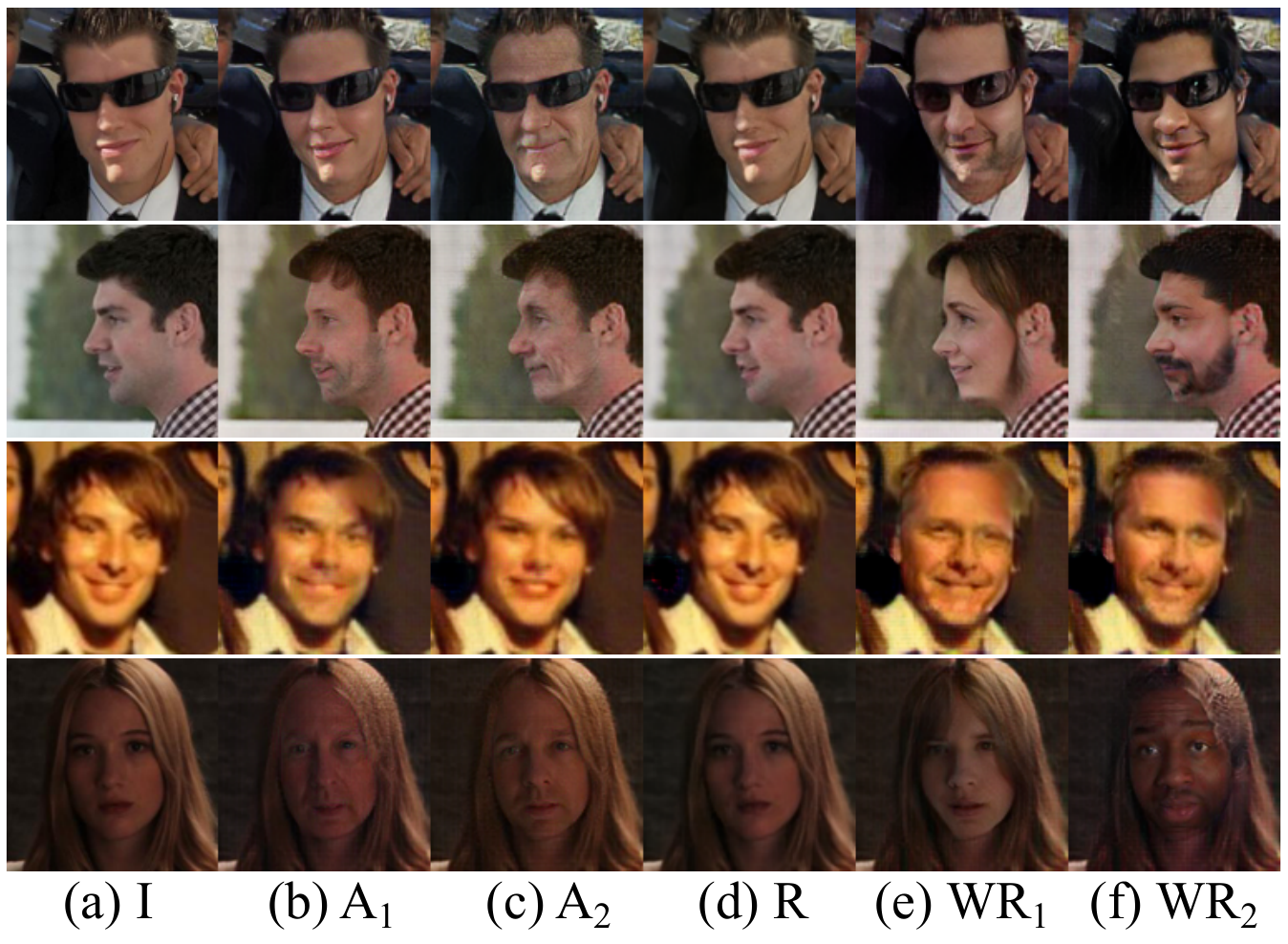}
}{
\caption{Hard cases on CASIA. See Fig.~\ref{fig:ffhq_and_lfw} caption for key.}
\label{fig:casia_hard}
}
\ffigbox{
\includegraphics[width=\linewidth]{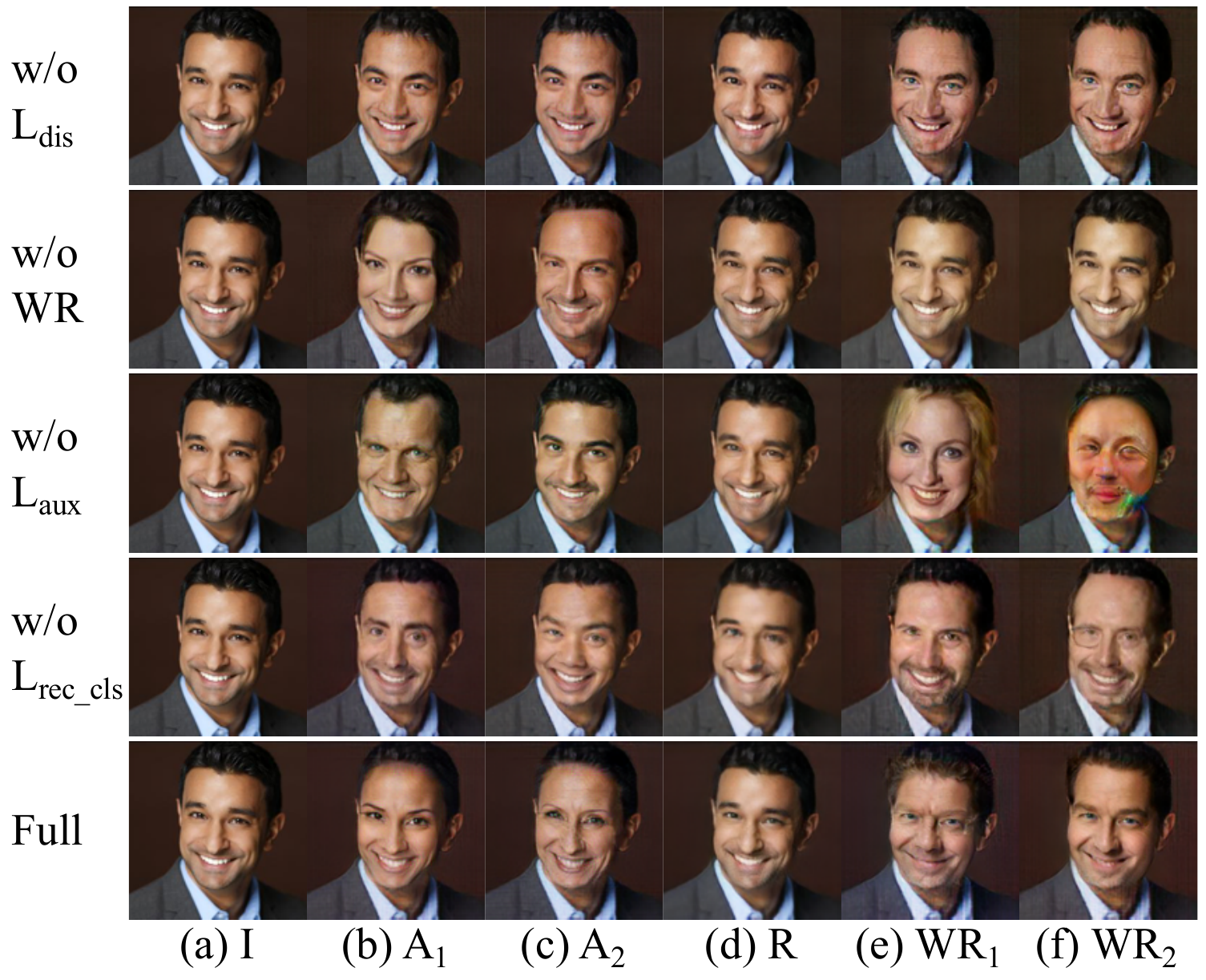}
}{
\caption{Typical failures of each ablation. See Fig.~\ref{fig:ffhq_and_lfw} caption for key.}
\label{fig:ablation}
}
\end{floatrow}
\end{figure}



\subsection{Generalization and difficult cases}
Fig.~\ref{fig:ffhq_and_lfw} shows generalization results on FFHQ and LFW using our model trained on CASIA. Without any fine-tuning, our model achieves good generalization performance on both the high quality FFHQ dataset and the LFW dataset where resolution is usually lower. 

Fig.~\ref{fig:casia_hard} shows hard-case qualitative results on CASIA. Our method works well even if the faces are with occlusions (sunglasses), with extreme poses, vague, under dim light, etc. We provide more qualitative results in supp.

\subsection{Applying CV algorithms on transformed faces}\label{sec:cvalgorithm}
Unlike most traditional anonymization algorithms~\cite{butler15,ryoo2017privacy}, our choice of achieving photo-realism on the (de)anonymizations makes it possible to apply existing computer vision algorithms directly on the transformed faces. To demonstrate this, we apply an off-the-shelf MTCNN~\cite{MTCNN} face bounding box and keypoint detector on the transformed faces. Qualitative detection results (see supp Fig.~5) are good. Quantitatively, although we do not have the ground truth annotations for transformed faces, we observe that our (de)anonymizations mostly do not change the head/keypoints' positions from the input faces so we can compare the detection results between the input faces and the transformed faces. Results are shown in Table~\ref{table:cv_algorithm}, which shows that a face detection algorithm trained on real images performs accurately on our transformed faces.

\subsection{Ablation studies} \label{subsec:ablation}
Finally, we evaluate the contribution of each component and loss in our model.  Here, original image ($I$), anonymized face with two different passwords ($A_{1,2}$), recovered face with correct inverse password ($R$), and recovered faces with wrong passwords ($\WR_{1,2}$):

\noindent\textbf{w/o $\mathbf{\LL_{dis}}$}: We remove feature dissimilarity loss on $(A_1, A_2)$ and $(\WR_1, \WR_2)$.

\noindent\textbf{w/o $\mathbf{\WR}$}: We do not explicitly train to produce wrong reconstructions.

\noindent\textbf{w/o $\mathbf{\LL_{aux}}$}: We remove the password-predicting auxiliary network $Q$, but still embed the passwords.

\noindent\textbf{w/o $\mathbf{\LL_{rec\_cls}}$}: We remove the face classification loss on the reconstruction.

Fig.~\ref{fig:ablation} shows the typical drawbacks of each ablation model. 
w/o $\LL_{dis}$ shows that $\LL_{dis}$ is necessary to achieve semantic-level multimodality on both anonymization and wrong reconstruction. 
w/o $\WR$ shows that without training for wrong reconstructions, the transformer fails to conceal identities when given incorrect passwords.
w/o $\LL_{aux}$ verifies the importance of the auxiliary network, which helps improve photo-realism and we also observe it helps with multimodality. 
Without $\LL_{rec\_cls}$, the reconstruction quality suffers because of unbalanced losses.

\begin{table}[t!]
\centering
\scriptsize
\begin{tabular}{lccc}
\hline
Avg spatial coordinate difference & CASIA & LFW & FFHQ  \\ 
\hline
Bounding boxes & 1.81 & 1.62 & 1.91\\
Keypoints & 0.94 & 0.76 & 0.89\\
\hline
\end{tabular}
\caption{Average pixel difference in detected coordinates of face bounding boxes and 5 keypoints between transformed faces ($A, R, \WR$) and input face ($I$).} \label{table:cv_algorithm}
\end{table}

\section{Discussion}
We presented a novel privacy-preserving face identity transformer with a password embedding scheme, multimodal identity change, and a multi-task learning objective.
We feel that this paper has shown the promise of password-conditioned face anonymization and deanonymization to address the privacy versus accessibility tradeoff.  
Although relatively rare, we sometimes notice artifacts that look similar to general GAN artifacts. They tend to arise due to the difficulty of image generation itself -- we believe they can be greatly reduced with more advances in image synthesis research, which can be (orthogonally) plugged into our system.

\paragraph{Acknowledgements.}
This work was supported in part by NSF IIS-1812850, NSF IIS-1812943, NSF CNS-1814985, NSF CAREER IIS-1751206, AWS ML Research Award, and Google Cloud Platform research credits. We thank Jason Ren, UC Davis labmates, and the reviewers for constructive discussions.

%
%
\bibliographystyle{splncs04}
\bibliography{egbib}

\begin{thebibliography}{10}
\providecommand{\url}[1]{\texttt{#1}}
\providecommand{\urlprefix}{URL }
\providecommand{\doi}[1]{https://doi.org/#1}

\bibitem{abadi2016privacy}
Abadi, M., Chu, A., Goodfellow, I., McMahan, H.B., Mironov, I., Talwar, K.,
  Zhang, L.: Deep learning with differential privacy. In: CCS (2016)

\bibitem{IPGAN}
Bao, J., Chen, D., Wen, F., Li, H., Hua, G.: Towards open-set identity
  preserving face synthesis. In: CVPR (2018)

\bibitem{butler15}
Butler, D.J., Huang, J., Roesner, F., Cakmak, M.: The privacy-utility tradeoff
  for remotely teleoperated robots. In: ICHRI (2015)

\bibitem{vggface2}
Cao, Q., Shen, L., Xie, W., Parkhi, O.M., Zisserman, A.: Vggface2: A dataset
  for recognising faces across pose and age. In: FG (2018)

\bibitem{cao2018hashgan}
Cao, Y., Liu, B., Long, M., Wang, J.: Hashgan: Deep learning to hash with pair
  conditional wasserstein gan. In: CVPR (2018)

\bibitem{chen2017semi}
Chen, J., Wu, J., Konrad, J., Ishwar, P.: Semi-coupled two-stream fusion
  convnets for action recognition at extremely low resolutions. In: WACV (2017)

\bibitem{infogan}
Chen, X., Duan, Y., Houthooft, R., Schulman, J., Sutskever, I., Abbeel, P.:
  Infogan: Interpretable representation learning by information maximizing
  generative adversarial nets. In: NeurIPS (2016)

\bibitem{erkin2009privacy}
Erkin, Z., Franz, M., Guajardo, J., Katzenbeisser, S., Lagendijk, I., Toft, T.:
  Privacy-preserving face recognition. In: PETS (2009)

\bibitem{gilad2016cryptonets}
Gilad-Bachrach, R., Dowlin, N., Laine, K., Lauter, K., Naehrig, M., Wernsing,
  J.: Cryptonets: Applying neural networks to encrypted data with high
  throughput and accuracy. In: ICML (2016)

\bibitem{GAN}
Goodfellow, I., Pouget-Abadie, J., Mirza, M., Xu, B., Warde-Farley, D., Ozair,
  S., Courville, A., Bengio, Y.: Generative adversarial nets. In: NeurIPS
  (2014)

\bibitem{LFW2}
Huang, G.B., Mattar, M., Berg, T., Learned-Miller, E.: Labeled faces in the
  wild: A database forstudying face recognition in unconstrained environments.
  In: Workshop on faces in `Real-Life' Images (2008)

\bibitem{pix2pix}
Isola, P., Zhu, J.Y., Zhou, T., Efros, A.A.: Image-to-image translation with
  conditional adversarial networks. In: CVPR (2017)

\bibitem{karras2018style}
Karras, T., Laine, S., Aila, T.: A style-based generator architecture for
  generative adversarial networks. arXiv:1812.04948  (2018)

\bibitem{adam}
Kingma, D.P., Ba, J.: Adam: A method for stochastic optimization.
  arXiv:1412.6980  (2014)

\bibitem{larsen2015autoencoding}
Larsen, A.B.L., S{\o}nderby, S.K., Larochelle, H., Winther, O.: Autoencoding
  beyond pixels using a learned similarity metric. arXiv:1512.09300  (2015)

\bibitem{Anonymousnet}
Li, T., Lin, L.: Anonymousnet: Natural face de-identification with measurable
  privacy. In: CVPR Workshops (2019)

\bibitem{sphereface}
Liu, W., Wen, Y., Yu, Z., Li, M., Raj, B., Song, L.: Sphereface: Deep
  hypersphere embedding for face recognition. In: CVPR (2017)

\bibitem{lsgan}
Mao, X., Li, Q., Xie, H., Lau, R.Y., Wang, Z., Paul~Smolley, S.: Least squares
  generative adversarial networks. In: ICCV (2017)

\bibitem{Mathieu_et_al}
Mathieu, M., Couprie, C., LeCun, Y.: Deep multi-scale video prediction beyond
  mean square error. arXiv:1511.05440  (2015)

\bibitem{mirza2014conditional}
Mirza, M., Osindero, S.: Conditional generative adversarial nets.
  arXiv:1411.1784  (2014)

\bibitem{resize_conv}
Odena, A., Dumoulin, V., Olah, C.: Deconvolution and checkerboard artifacts.
  Distill  (2016). \doi{10.23915/distill.00003},
  \url{http://distill.pub/2016/deconv-checkerboard}

\bibitem{perarnau2016invertible}
Perarnau, G., Van De~Weijer, J., Raducanu, B., {\'A}lvarez, J.M.: Invertible
  conditional gans for image editing. arXiv:1611.06355  (2016)

\bibitem{ganimation}
Pumarola, A., Agudo, A., Martinez, A.M., Sanfeliu, A., Moreno-Noguer, F.:
  Ganimation: Anatomically-aware facial animation from a single image. In: ECCV
  (2018)

\bibitem{reed2016generative}
Reed, S., Akata, Z., Yan, X., Logeswaran, L., Schiele, B., Lee, H.: Generative
  adversarial text to image synthesis. arXiv:1605.05396  (2016)

\bibitem{Regmi_2018_CVPR}
Regmi, K., Borji, A.: Cross-view image synthesis using conditional gans. In:
  CVPR (2018)

\bibitem{jason}
Ren, Z., Lee, Y.J., Ryoo, M.S.: Learning to anonymize faces for privacy
  preserving action detection. In: ECCV (2018)

\bibitem{rivest1978method}
Rivest, R.L., Shamir, A., Adleman, L.: A method for obtaining digital
  signatures and public-key cryptosystems. Communications of the ACM  (1978)

\bibitem{ryoo2018extreme}
Ryoo, M.S., Kim, K., Yang, H.J.: Extreme low resolution activity recognition
  with multi-siamese embedding learning. In: AAAI (2018)

\bibitem{ryoo2017privacy}
Ryoo, M.S., Rothrock, B., Fleming, C., Yang, H.J.: Privacy-preserving human
  activity recognition from extreme low resolution. In: AAAI (2017)

\bibitem{shen2017learning}
Shen, W., Liu, R.: Learning residual images for face attribute manipulation.
  In: CVPR (2017)

\bibitem{inpainting}
Sun, Q., Ma, L., Joon~Oh, S., Van~Gool, L., Schiele, B., Fritz, M.: Natural and
  effective obfuscation by head inpainting. In: CVPR (2018)

\bibitem{wang2016studying}
Wang, Z., Chang, S., Yang, Y., Liu, D., Huang, T.S.: Studying very low
  resolution recognition using deep networks. In: CVPR (2016)

\bibitem{DSSIM}
Wang, Z., Bovik, A.C., Sheikh, H.R., Simoncelli, E.P., et~al.: Image quality
  assessment: from error visibility to structural similarity. IEEE TIP
  \textbf{13}(4),  600--612 (2004)

\bibitem{CASIA}
Yi, D., Lei, Z., Liao, S., Li, S.Z.: Learning face representation from scratch.
  arXiv:1411.7923  (2014)

\bibitem{yonetani2017privacy}
Yonetani, R., Naresh~Boddeti, V., Kitani, K.M., Sato, Y.: Privacy-preserving
  visual learning using doubly permuted homomorphic encryption. In: ICCV (2017)

\bibitem{zhang2017stackgan}
Zhang, H., Xu, T., Li, H., Zhang, S., Wang, X., Huang, X., Metaxas, D.N.:
  Stackgan: Text to photo-realistic image synthesis with stacked generative
  adversarial networks. In: ICCV (2017)

\bibitem{MTCNN}
Zhang, K., Zhang, Z., Li, Z., Qiao, Y.: Joint face detection and alignment
  using multitask cascaded convolutional networks. IEEE Signal Processing
  Letters  \textbf{23}(10),  1499--1503 (2016)

\bibitem{lpips}
Zhang, R., Isola, P., Efros, A.A., Shechtman, E., Wang, O.: The unreasonable
  effectiveness of deep features as a perceptual metric. In: CVPR (2018)

\bibitem{cyclegan}
Zhu, J.Y., Park, T., Isola, P., Efros, A.A.: Unpaired image-to-image
  translation using cycle-consistent adversarial networks. In: ICCV (2017)

\bibitem{multimodal}
Zhu, J.Y., Zhang, R., Pathak, D., Darrell, T., Efros, A.A., Wang, O.,
  Shechtman, E.: Toward multimodal image-to-image translation. In: NeurIPS
  (2017)

\bibitem{zhu2017your}
Zhu, S., Urtasun, R., Fidler, S., Lin, D., Change~Loy, C.: Be your own prada:
  Fashion synthesis with structural coherence. In: ICCV (2017)

\end{thebibliography}

\clearpage
\appendix
\section*{Appendices}
\addcontentsline{toc}{section}{Appendices}
\renewcommand{\thesubsection}{\Alph{subsection}}

\begin{wrapfigure}{r}{0.5\textwidth}
\centering
\includegraphics[width=\linewidth]{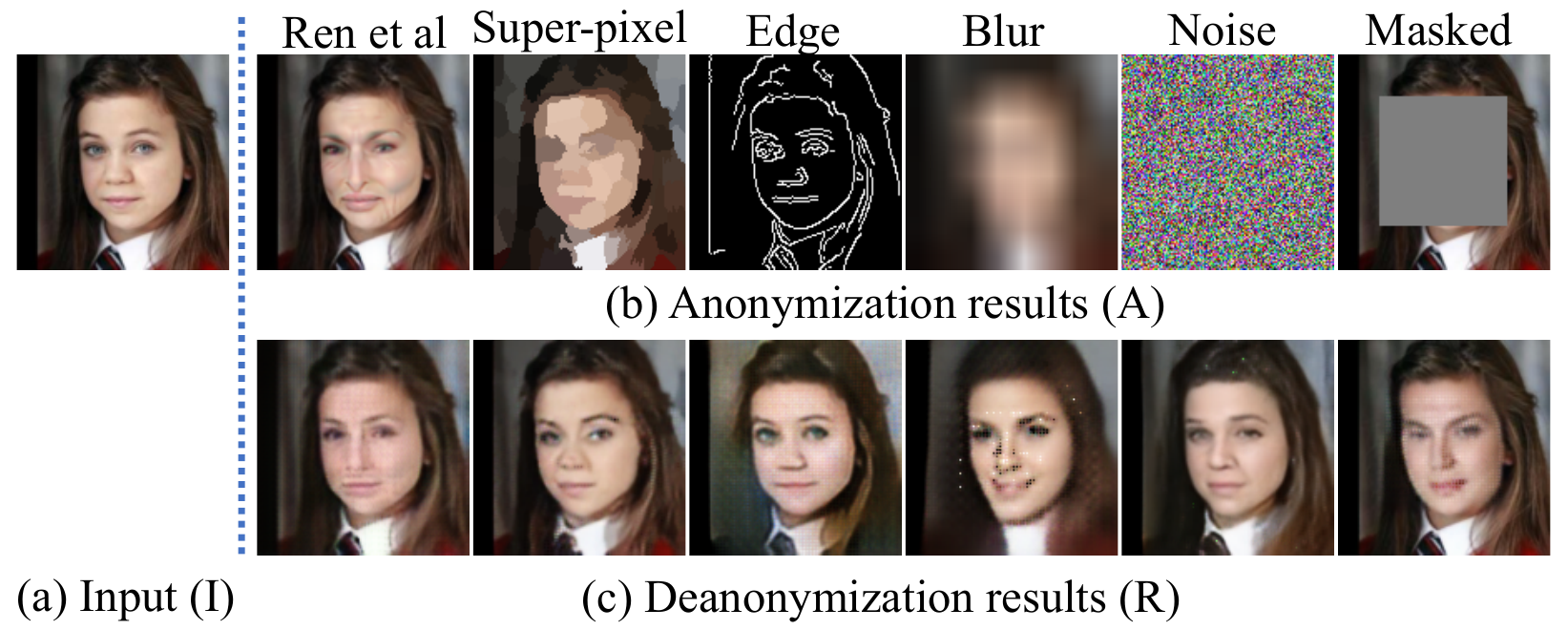}
\caption{\footnotesize{Baselines. Super-pixel, Edge, Blur, Noise, Masked sacrifice photo-realism for anonymization.}}
\label{fig:baseline}
\end{wrapfigure}

\subsection{Additional details}
Fig.~\ref{fig:baseline} shows a qualitative example of the baselines and their anonymizations/deanonymizations.

We use batch normalization and our transformer $T$ is based on the 9-block Resnet generator from \cite{cyclegan}. We also replace the transformer $T$'s fractionally-strided convolution layers with the resize-convolution layers in~\cite{resize_conv} to alleviate checkerboard artifacts. 

For auxiliary network $Q$ that predicts the embedded passwords, since there are a total of $2^N$ passwords, it is not ideal to have a $2^N$-way classifier when $N$ is large. Instead, we set up $N/4$ 16-way classifiers, with each classifier responsible for classifying its corresponding 4 bits into $2^4$ classes.

Let $p_i \in \{0, ..., 15\}$ denote a 4-bit chunk of $p$ and $\hat{p}_i$  denote the chunk predicted by $Q$. $Q(I, T_pI) = (f_1, …, f_{N/4})$, where $f_i$ is a 16-dim vector (logit). $Prob(\hat{p}_i = j) = Softmax(f_i)_j$. 
\begin{align}
\mathcal{L}_{aux}(T, Q) = -\sum_{i=1}^{N/4} \log( Prob(\hat{p}_i = p_i ) ).	
\end{align}

For $Q$'s architecture, we modify PatchGAN by switching the last convolutional layer to an average pooling layer followed by $N/4$ parallel fully-connected layers that predict the passwords. 

The face recognition model $F$ (SphereFace \cite{sphereface}) is trained on aligned and cropped faces, so during training, we use the same manner of aligning face by facial landmarks before inputting any faces to $F$ as in~\cite{jason}. The facial landmarks are detected by MTCNN~\cite{MTCNN}. For the VGGFace2~\cite{vggface2} face recognition model, we follow the same setting as the original paper: We use MTCNN~\cite{MTCNN} for face detection. The bounding boxes are then expanded by a factor 1.3x to include the whole head, which are used as network inputs.

All networks in our architecture were trained from scratch with a learning rate of 0.0001 for 15 epochs except the pre-trained face recognition model which used a learning rate of 0.00001. We use Adam solver~\cite{adam} and a total batch size of 48 on 4 GPUs.

For the AMT photo-realism test, we do not include the synthesized images in which a man's face is with hair that obviously belongs to a woman; in such cases, Turkers may attribute fakeness to prior experience (it is uncommon to see a man with a woman's hairstyle) rather than photo-realism. 
This could be resolved by training separate face identity transformers for each gender. 

\subsection{Discussion on reverse engineering}

Threat models to our model are either white-box (have complete knowledge of $T$) or black-box (get input-output pairs from $T$).

Theoretically speaking, assuming all desiderata are achieved:

\begin{itemize}
	\item Since every password leads to a unique photorealistic identity, without prior knowledge, a brute-force adversary cannot decide which one is correct.
	\item Adversaries $\mathcal{V}$ in the form of $\mathcal{V}_1(T_pI) = \hat{p}$ or $\mathcal{V}_2(T_pI) = \hat{I}$ won’t work. We can use any password $p’$ to anonymize and deanonymize $T_pI$ and still get $T_pI$, but in this case $\mathcal{V}_1$ should output $-p’$:  
\begin{align}
\forall p’, \hat{p} = \mathcal{V}_1(T_pI) =  \mathcal{V}_1(T_{-p’}(T_{p’}T_pI)) = -\hat{p}’, contradict!
\end{align}
Similar argument applies to $\mathcal{V}_2$. Note that different from adversaries, our auxiliary network $Q$ also takes the original face as input.
\end{itemize}

In pracitce, due to existing artifacts in GANs, the desiderata are not perfectly achieved. And thus our current model cannot achieve this theoretical robustness against adversaries. We believe 1) Orthogonally plugging in better image synthesizing techniques; 2) Explicitly introducing robustness against adversaries are the future directions to pit against reverse engineering.

\subsection{Discussion on wrong reconstruction better hides identity}

Both qualitative results and AMT studies show that Wrongly Recovered faces ($\WR$s) better hide identities. We believe this is happening because:
\begin{itemize}
	\item $\WR$ has less constraints to satisfy compared to Anonymized faces ($A$) in our loss formulation. Our training process could lead $\WR$ to become more optimized for the face classification loss as it does not need to care about the reconstruction loss, while $A$ does need to be optimized to allow reconstruction of Recovered faces ($R$).
	\item $\WR$ is a result of two transformations from the input face rather than one. while $A$ is a result of only a single transformation. More transformations lead to more identity changes (though we also notice more artifacts in $\WR$ than $A$).
\end{itemize}
Increasing the weight of the face classification loss $\mathcal{L}_{adv}$ applied to $A$ may make $A$ hide identity better.

\begin{figure}[t!]
\centering
\includegraphics[width=0.45\linewidth]{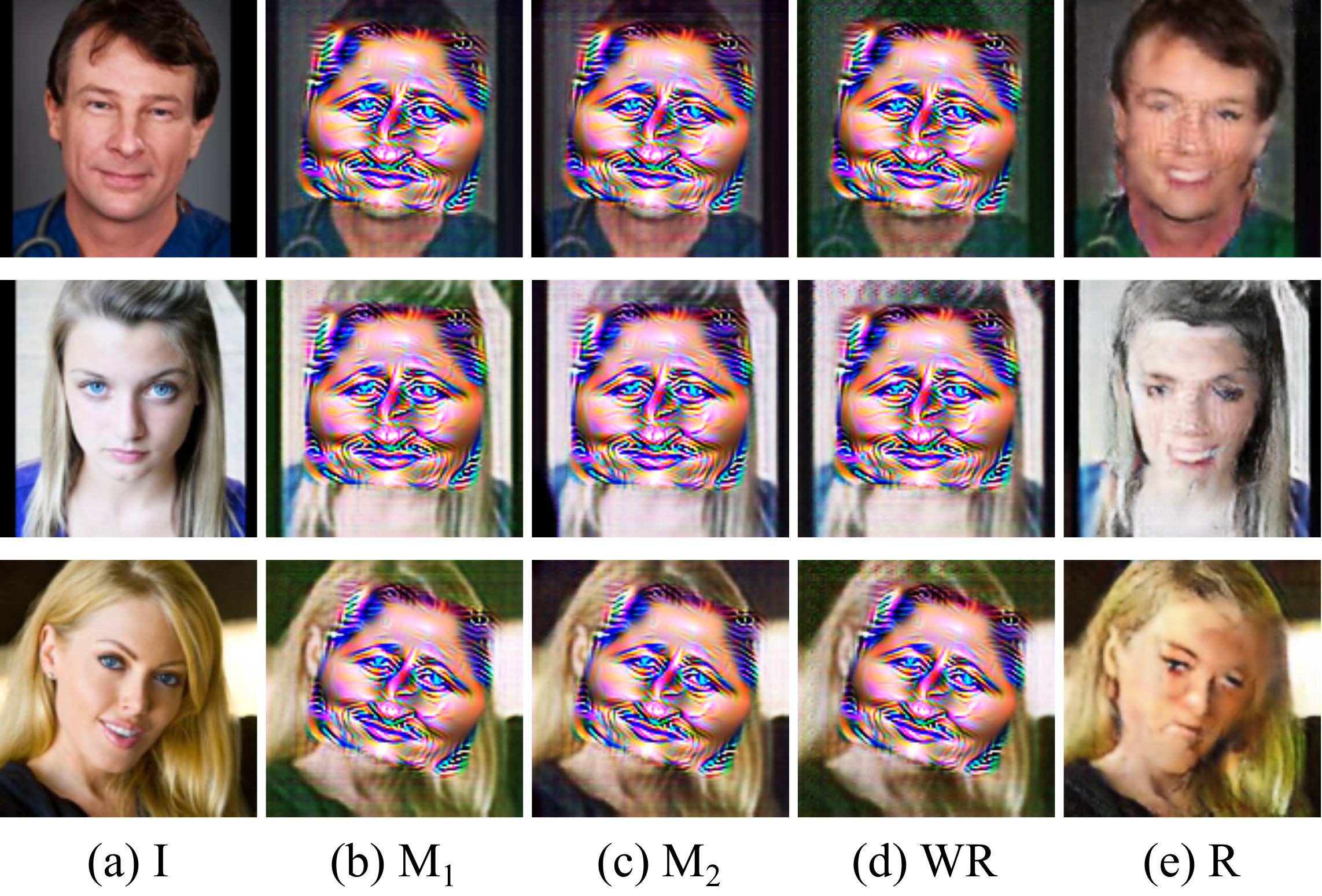}
   \caption{Ablation study on CASIA-WebFace trained with non-adversarial face classification loss Eq.~\ref{eq:non_adv} , which shows that this loss dominates the multi-task learning objective quickly, so adversarial training on face classification is necessary.}
\label{fig:no_FR_adv}
\end{figure}

\subsection{Why do we update the face classifier during training?}

This is an adversarial learning setting that makes the transformer more robust. During each generator’s stage, we train $T$ to make $T_pI$ have a different identity from $I$. During each discriminator’s stage, we train $F$ to correctly classify $I$ as well as classify $T_pI$ as $y_I$, i.e., see through the disguise of $T_pI$. $T$ and $F$ compete against each other so that our anonymization has certain robustness under the attack of finetuning $F$. We don’t want to disturb the pretraining of $F$ too much, so we set a much lower learning rate for $F$, see Sec.~1.

We also did an ablation study where the face flassifier $F$ is fixed during training in a non-adversarial training manner, i.e., we replace Eq.~10 in the main paper with:
\begin{align}
\LL_{non\_adv}(T) = & -\E_{(I,p)} \LL_{CE}(F(T_pI), y_I) \nonumber\\
 &- \E_{(I, p' \neq -p)} \LL_{CE}(F(T_{p'}T_pI), y_I)\label{eq:non_adv},
\end{align}

Fig.~\ref{fig:no_FR_adv} shows the common failure pattern: the anonymizations are no longer photorealistic but all have a common very fake face and reconstruction also suffers.
These results indicate that this setting does not work. As shown from the loss curve, the misclassification loss quickly turns into large magnitude and dominates the full objective. On the other hand, the adversarial training makes the misclassification loss not easily satisfied and not dominating.

%

\subsection{Additional results}

\begin{figure}[t]
\centering
\includegraphics[width=0.5\linewidth]{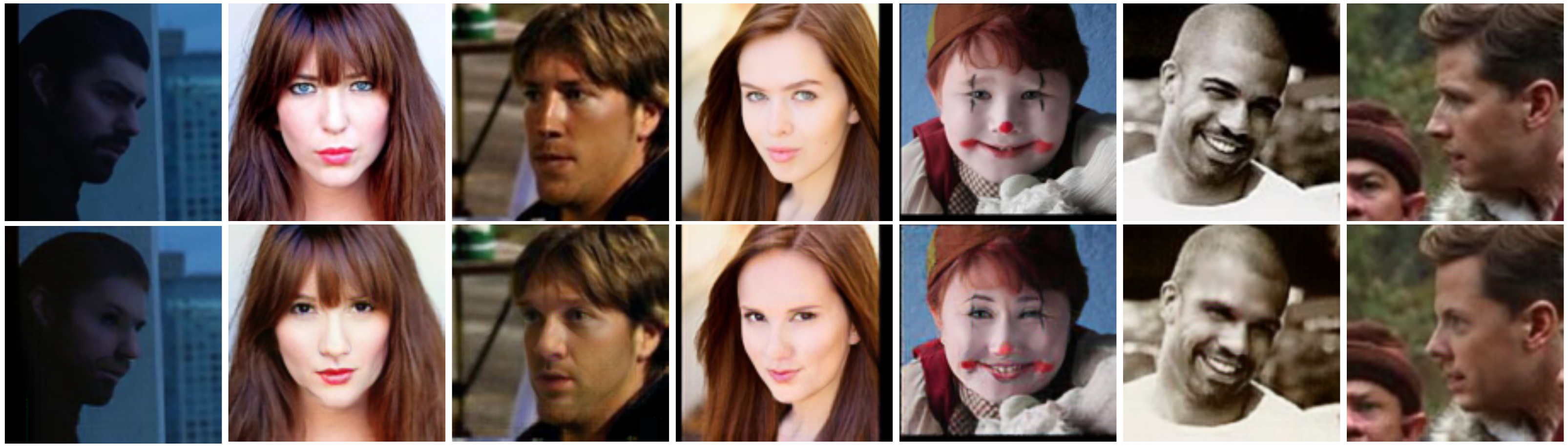}
  \caption{All pairs of inputs (top) \& anonymizations (bottom) turkers reported as same person. Our model still works to some extent.}\label{fig:error_pairs}
\end{figure}

In Fig.~\ref{fig:error_pairs}, we show all 7 out of 150 pairs (4.7\%) that turkers report as the input and anonymized faces belonging to the same person. Even though the turkers reported ``yes'', our transformer still works to some extent -- it changes color of skin/eyes, shape of eyes/nose/mouth/facial muscles. The same background and the same hair styles may have confused the turkers. In addition, they are mostly hard cases: dim light, side faces, heavy paints, and grayscale images. For these cases we do not have enough samples in the training set. If we collect more samples of these cases, we expect the model to perform better.

The quantitative reconstruction results on FFHQ~\cite{karras2018style} is 0.0602/0.0471/0.0509/ 0.0057 for LSIPS/SSIM/L1/L2, as a supplement for Table 2 in the main paper, which indicates that our transformer generalizes well on the deanonymization task on FFHQ, a dataset with plentiful variation in age, ethnicity and image background.

\begin{figure*}
\centering
\includegraphics[width=1.0\linewidth]{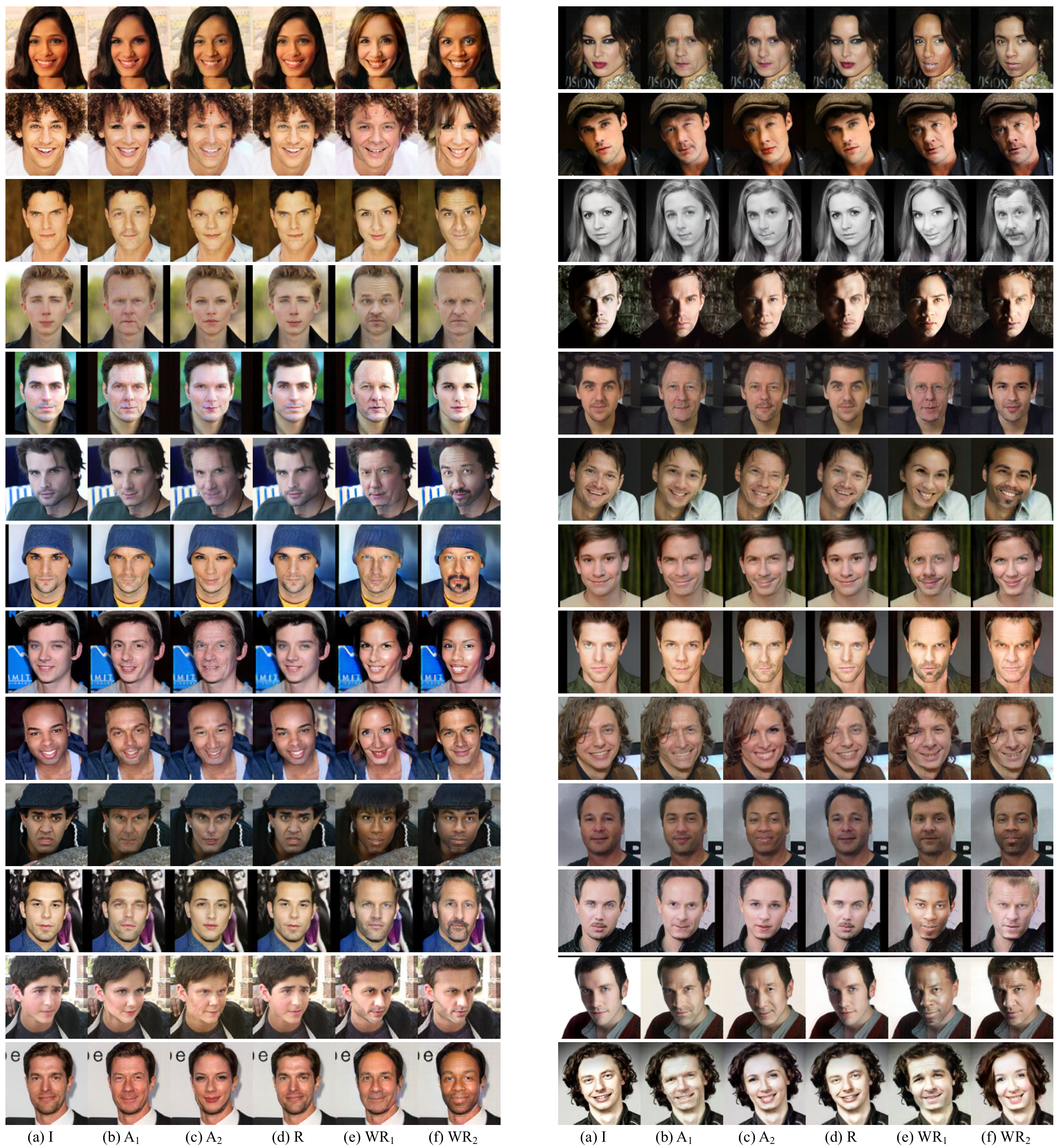}
   \caption{Additional qualitative results on CASIA. $I$: original image, $A_{1,2}$: anonymized faces conditioned on different passwords, $R/\WR_{1,2}$: recovered faces with correct/wrong passwords.}
\label{fig:add_qual_casia}
\end{figure*}

We show more qualitative results on CASIA~\cite{CASIA} in Fig.~\ref{fig:add_qual_casia}. For faces of different hair styles, poses, and ages, our model produces high-quality results.

\begin{figure}[htb!]
\centering
\subfloat[CASIA]{
\includegraphics[width=0.45\linewidth]{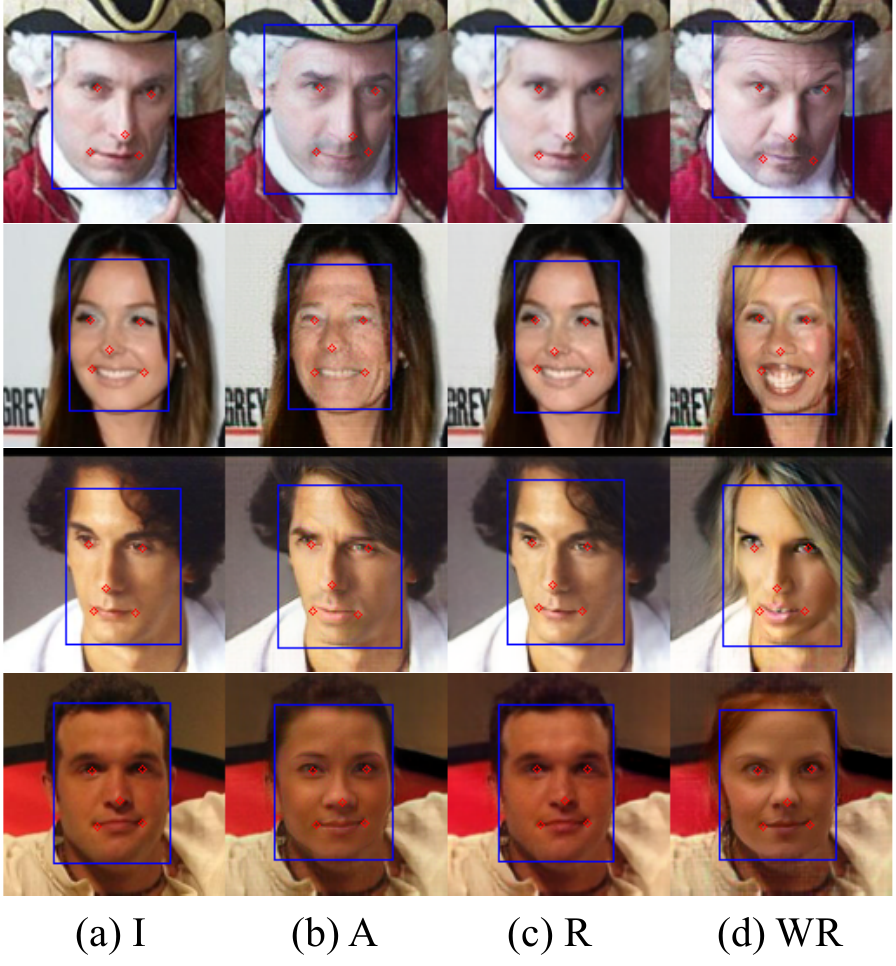}
}
\subfloat[FFHQ]{
\includegraphics[width=0.45\linewidth]{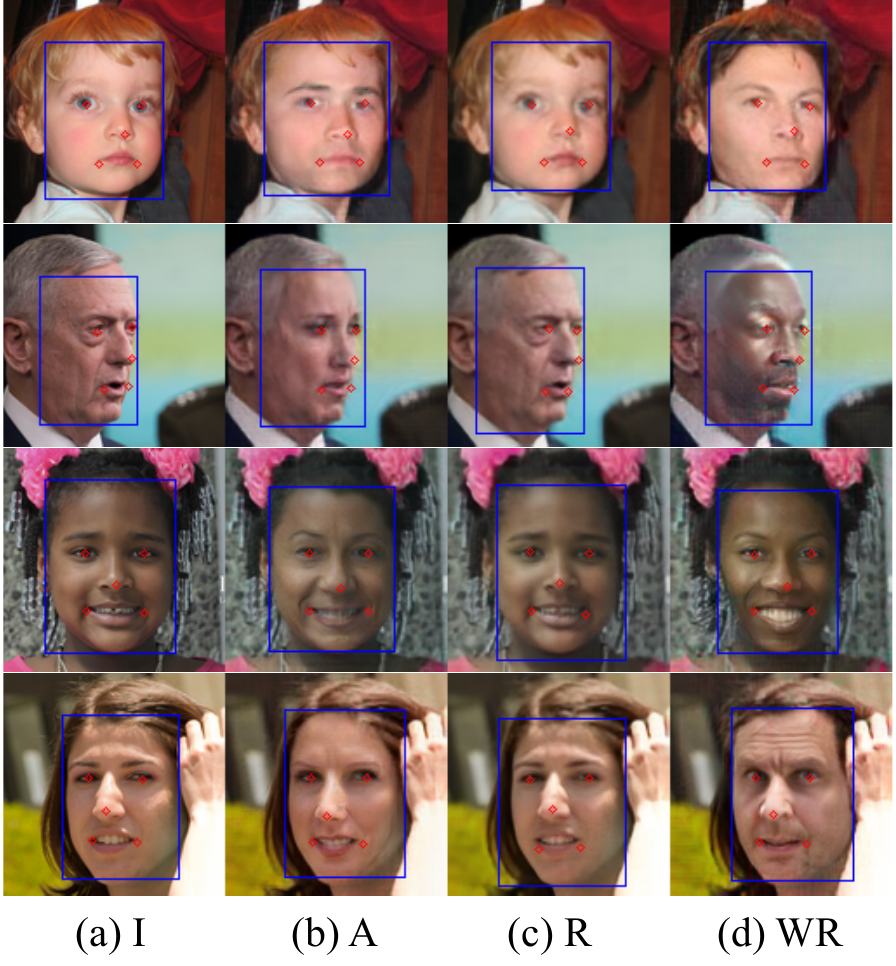}
}

\subfloat[LFW]{
\includegraphics[width=0.45\linewidth]{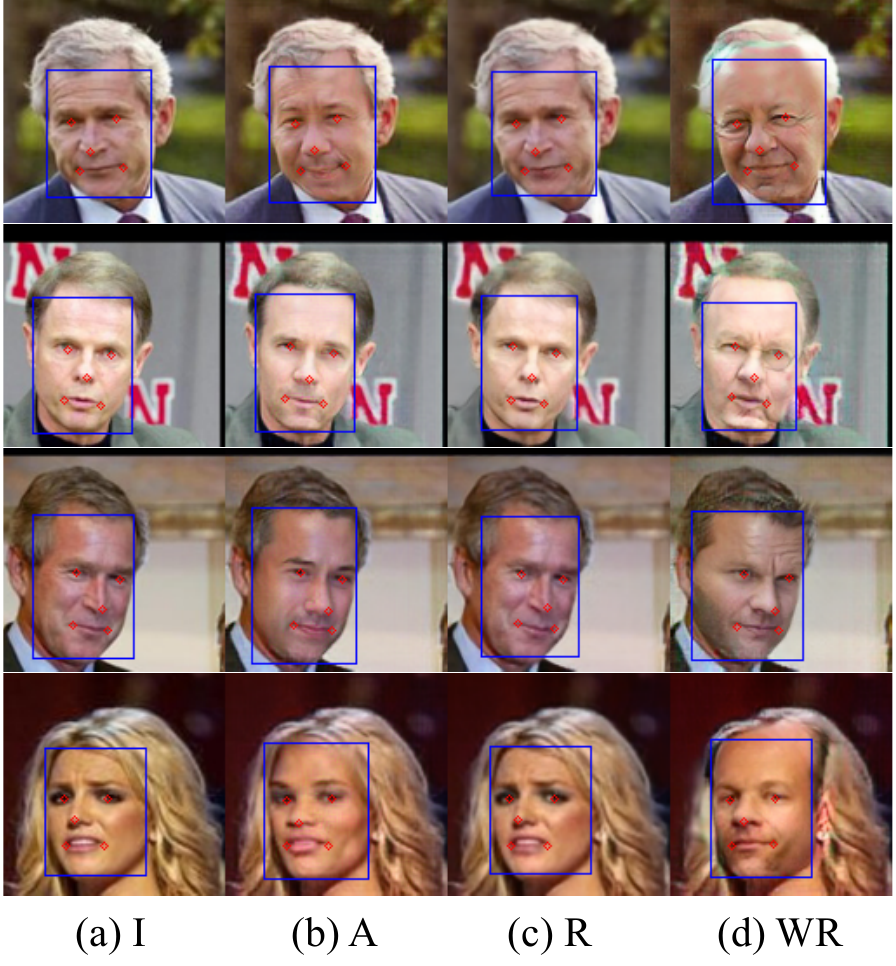}
}
\caption{Qualitative face detection results on transformed images. Photo-realism makes existing computer vision algorithms work on our transformed images directly.}
\label{fig:detection}
\end{figure}

Fig.~\ref{fig:detection} shows qualitative face detection results when applying an off-the-shelf face detector (MTCNN~\cite{MTCNN}) on the transformed images, see Table 3 in the main paper for quantitative results. The good performance demonstrates that normal computer vision algorithms developed on real images can be directly applied on our transformed faces, which is a great advantage over traditional face anonymization approaches.

\begin{figure*}[p]
\centering
\subfloat[Original image]{
\includegraphics[width=0.48\textwidth]{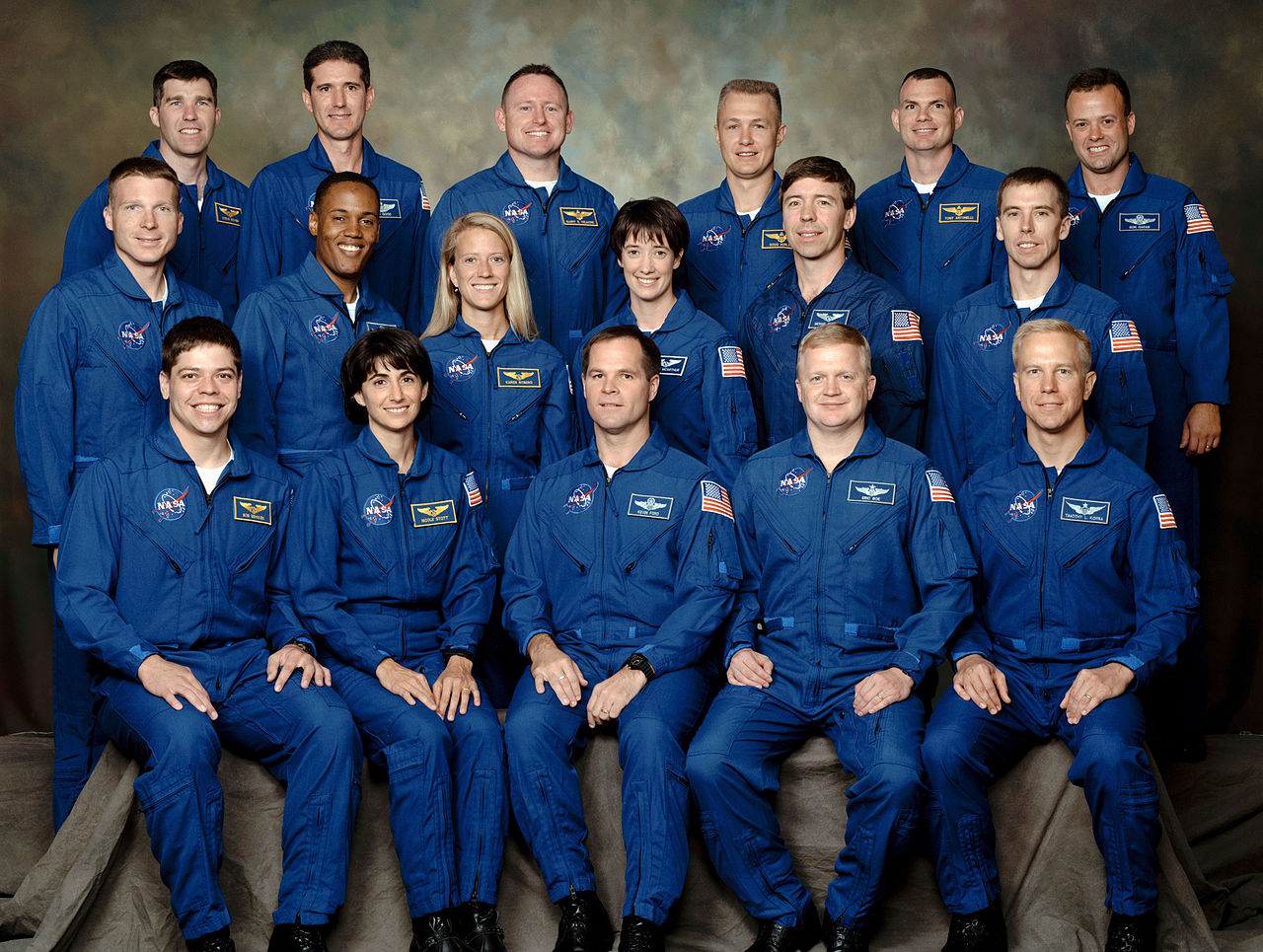}
} 
\subfloat[Anonymized image]{
\includegraphics[width=0.48\textwidth]{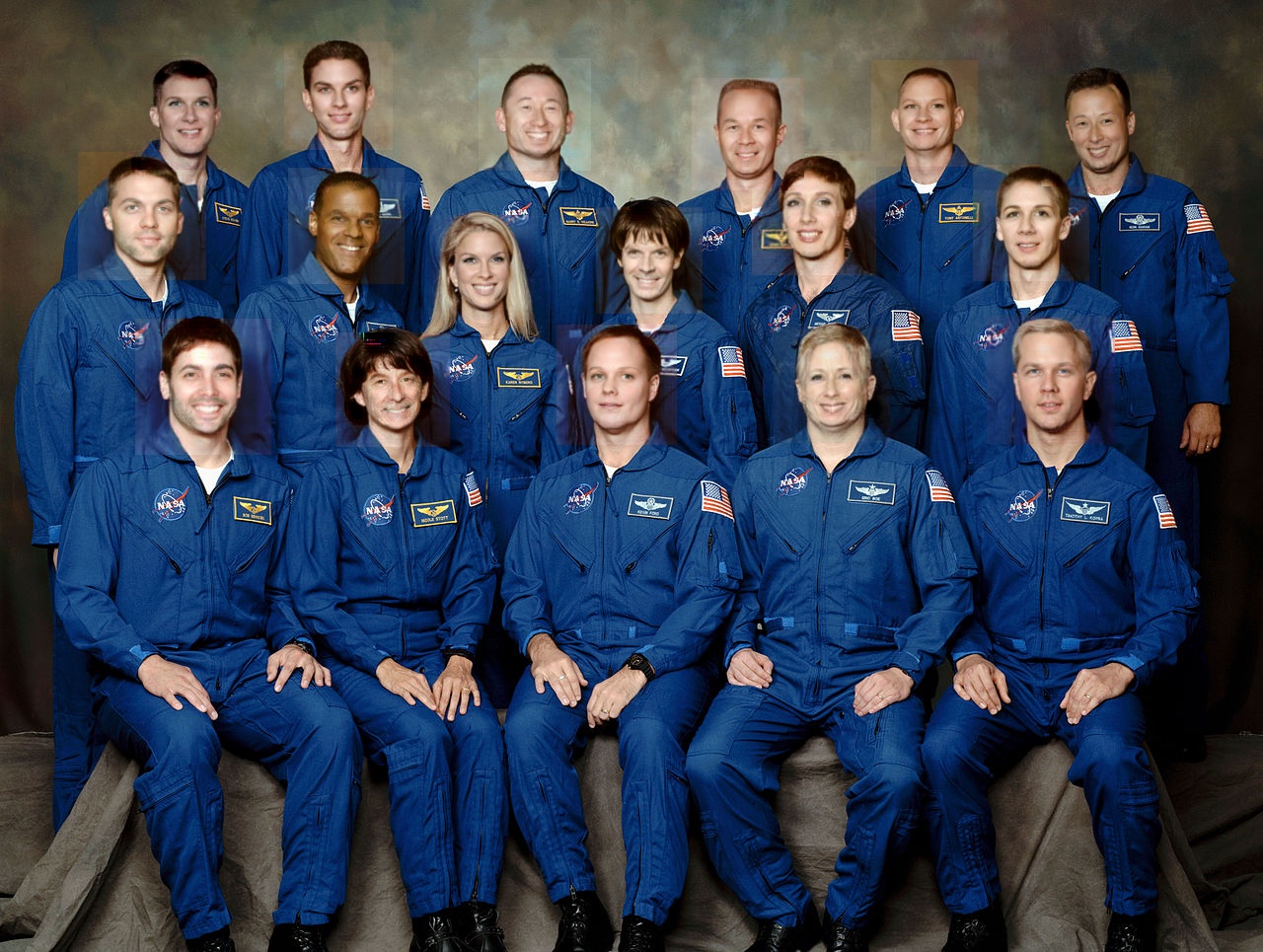}
}

\subfloat[Deanonymized image with correct password]{
\includegraphics[width=0.48\textwidth]{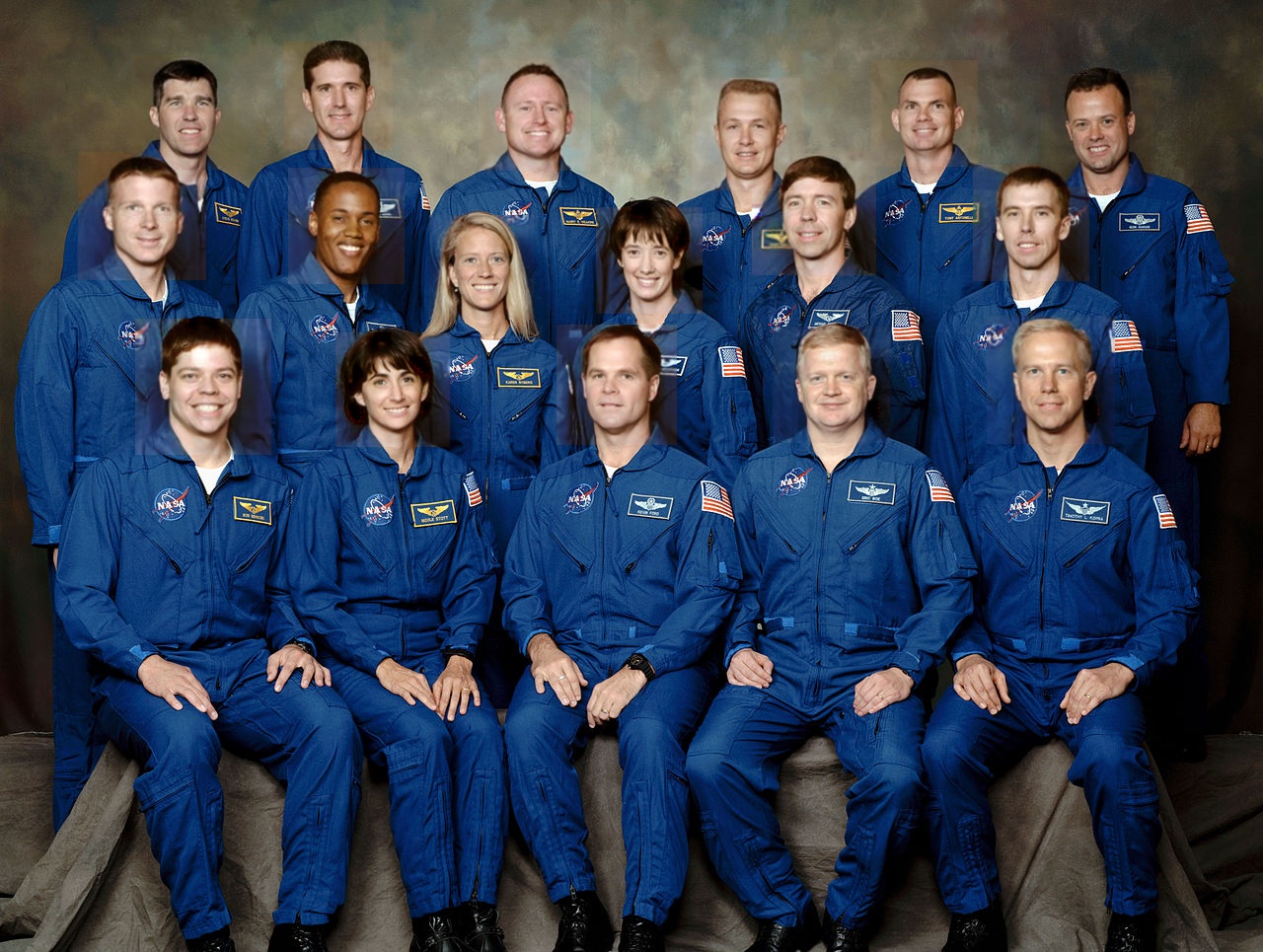}
}
\subfloat[Deanonymized image with wrong password]{
\includegraphics[width=0.48\textwidth]{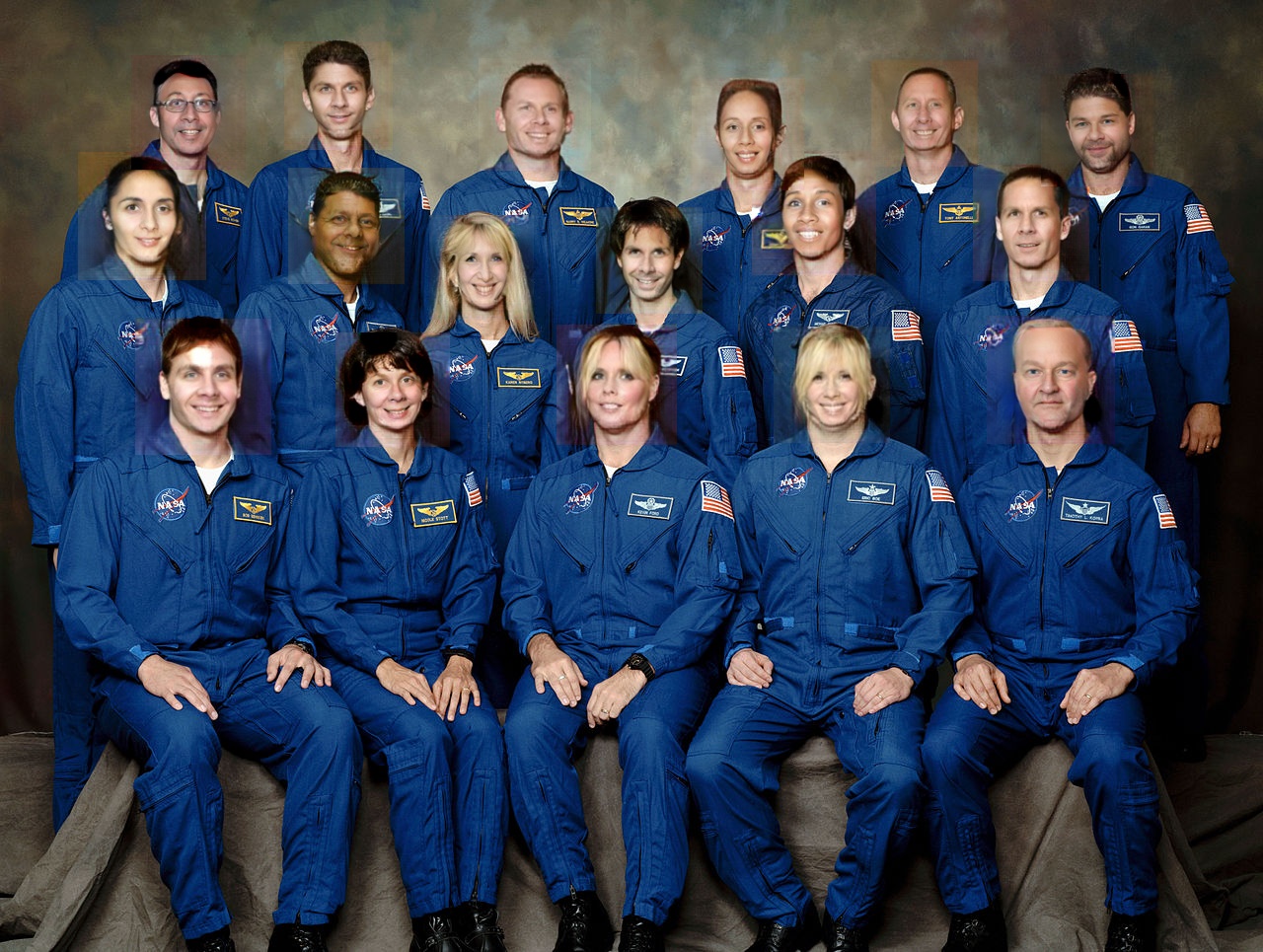}
}

\caption{Image in the wild example.}
\label{fig:wild}
\end{figure*}

\subsection{Image in the wild}
In Fig.~\ref{fig:wild}, we show that with the help of an off-the-shelf face detector, MTCNN~\cite{MTCNN}, our system works well on images in the wild. The anonymized and deanonymized face areas fit well into the original image. Please also check our uploaded video at \url{https://youtu.be/FrYmf-CL4yk}, which demonstrates that our model can be consistent in time.

\subsection{Further exploration of the password scheme}
We further investigate how our password scheme works and what the transformer learns. Since the 16-bit password space has a total of 65,536 different passwords, which is a very large space to explore, we trained an additional model with 8-bit password scheme for experiments in this section.

\begin{figure}[t!]
\subfloat[]{\includegraphics[width=.1\linewidth]{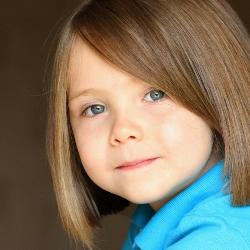}}
\subfloat[]{\includegraphics[width=.1\linewidth]{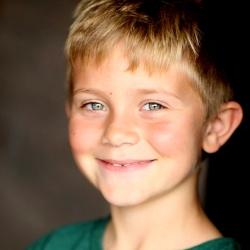}}
\subfloat[]{\includegraphics[width=.1\linewidth]{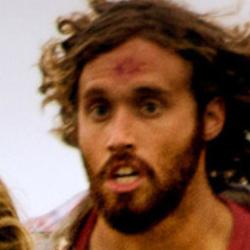}}
\caption{Original images. (a) and (b) are similar. (c) is more different from (a) and (b).}
\label{fig:reals}
\end{figure}

\begin{figure*}[b]
\centering
\includegraphics[width=1.0\linewidth]{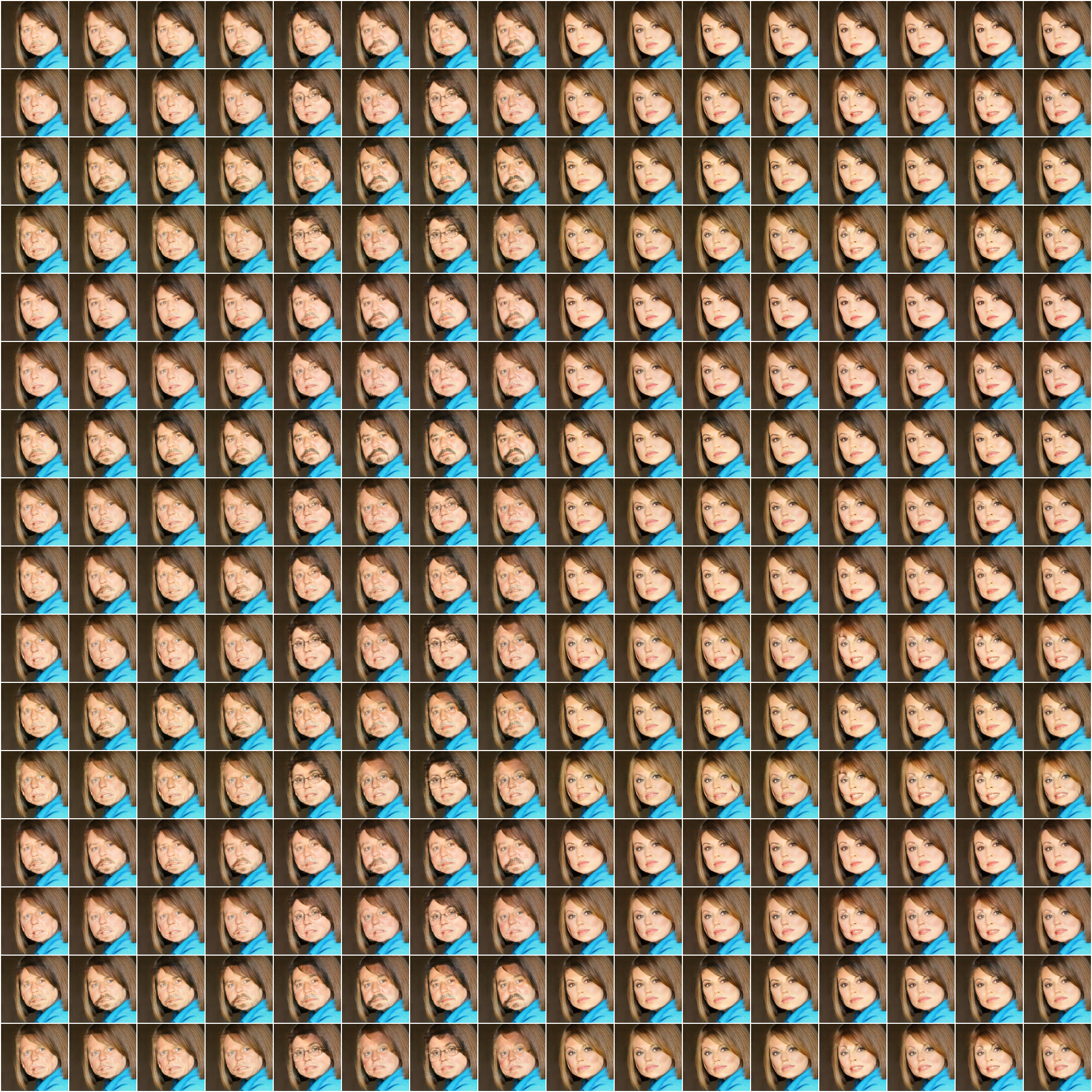}
   \caption{Modifications associated with all the passwords whose original face image is Fig.~\ref{fig:reals}(a).}
\label{fig:qualitative_a}
\end{figure*}

\begin{figure*}[p]
\centering
\includegraphics[width=1.0\linewidth]{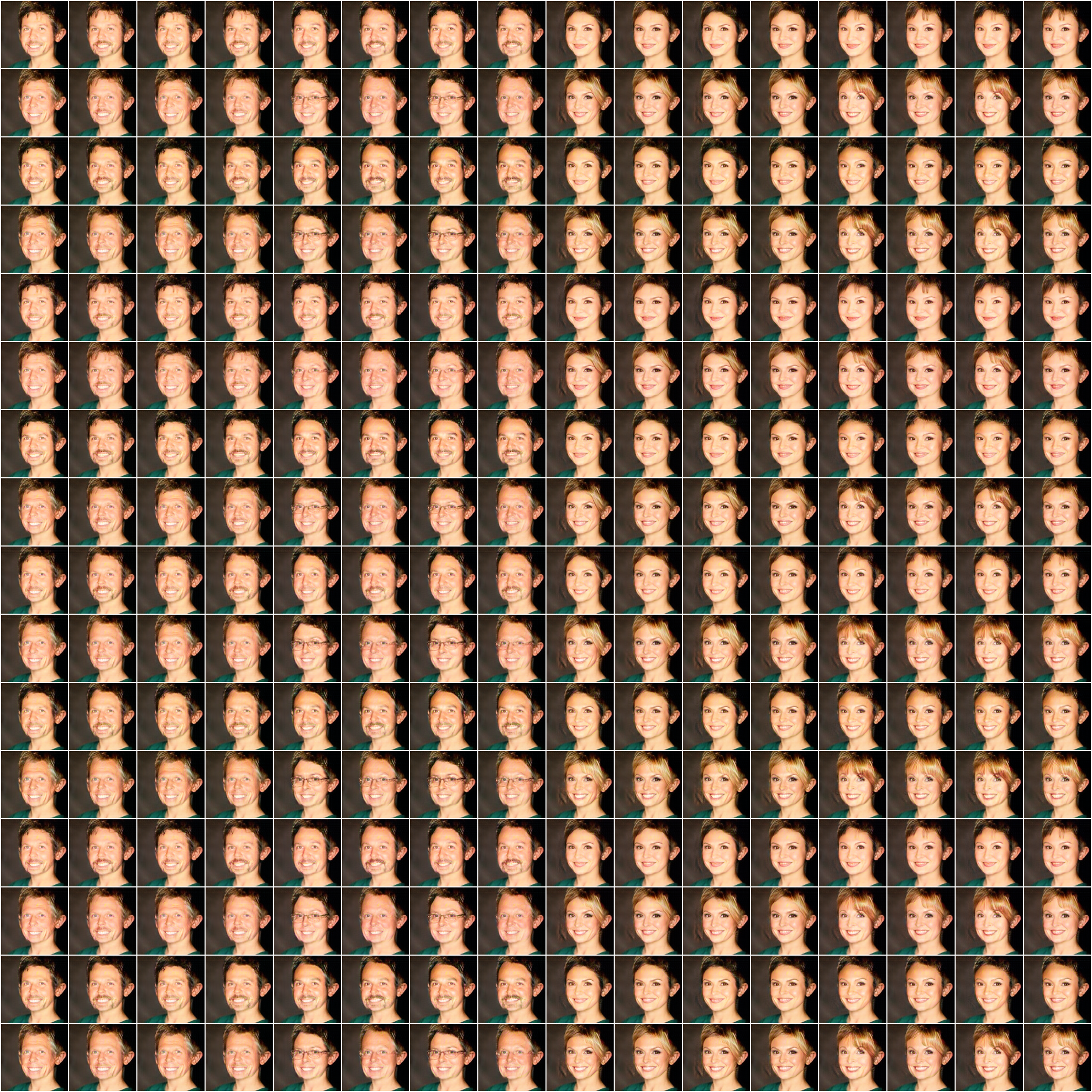}
   \caption{Modifications associated with all the passwords whose original face image is Fig.~\ref{fig:reals}(b).}
\label{fig:qualitative_b}
\end{figure*}

\begin{figure*}[p]
\centering
\includegraphics[width=1.0\linewidth]{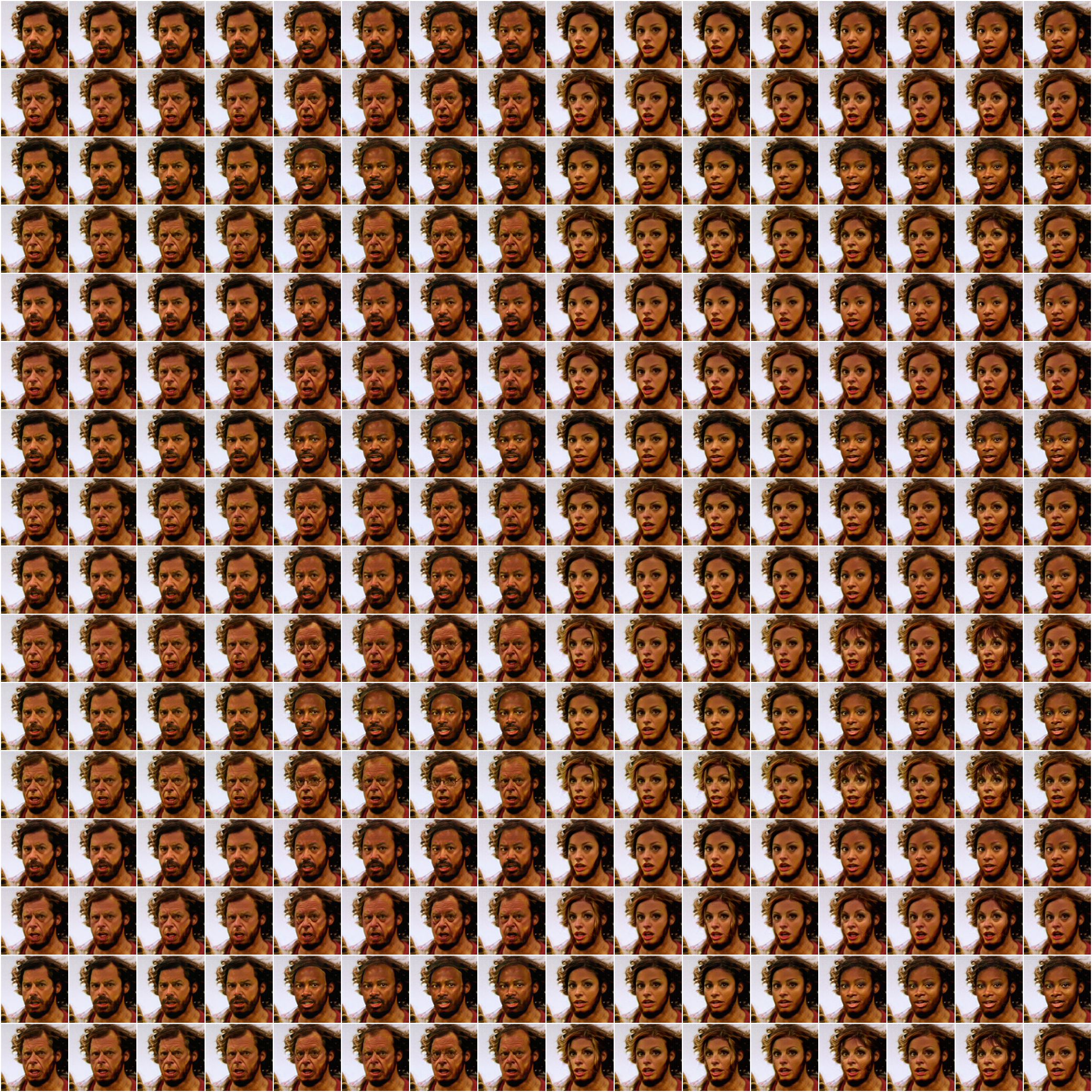}
   \caption{Modifications associated with all the passwords whose original face image is Fig.~\ref{fig:reals}(c).}
\label{fig:qualitative_c}
\end{figure*}



We show the modifications associated with all the passwords for the exemplar input images (Fig.~\ref{fig:reals}) in Fig.~\ref{fig:qualitative_a}, \ref{fig:qualitative_b}, \ref{fig:qualitative_c} respectively, where Fig.~\ref{fig:reals}(a) and Fig.~\ref{fig:reals}(b) are both children and Fig.~\ref{fig:reals}(c) is more different in age and appearance.

From the qualitative results we observe that similar original faces lead to similar modifications when given the same password. Interestingly, our transformer achieves gender equality -- half of the passwords transform to female identities and the remaining half transform to males regardless of the inputs' genders. 
And all the transformed faces satisfy our anonymization goal. These qualitative results also show that more diverse passwords lead to more diverse anonymized faces.

\end{document}